\documentclass[10pt,twocolumn,letterpaper]{article}

\usepackage{cvpr}
\usepackage{times}
\usepackage{epsfig}
\usepackage{graphicx}
\usepackage{amssymb}
\usepackage{balance}

\usepackage{amsmath,amsfonts}
\usepackage{algorithmic}
\usepackage{algorithm}
\usepackage{array}
\usepackage{epsfig}
\usepackage{subcaption}
\usepackage{float}
\usepackage{textcomp}
\usepackage{stfloats}
\usepackage{url}
\usepackage{verbatim}
\usepackage{graphicx}
\usepackage{booktabs}
\usepackage{adjustbox}
\usepackage{hyperref}
\usepackage{float}

\begin{document}

\title{Synthesizing Iris Images using Generative Adversarial Networks: \\ Survey and Comparative Analysis}

\author{Shivangi Yadav\\
Michigan State University\\
{\tt\small yadavshi@msu.edu}
\and
Arun Ross\\
Michigan State University\\
{\tt\small rossarun@msu.edu}
}

\maketitle
\thispagestyle{empty}

\begin{abstract}
   Biometric systems based on iris recognition are currently being used in border control applications and mobile devices. However, research in iris recognition is stymied by various factors such as limited datasets of bonafide irides and presentation attack instruments; restricted intra-class variations; and privacy concerns. Some of these issues can be mitigated by the use of synthetic iris data. In this paper, we present a comprehensive review of state-of-the-art GAN-based synthetic iris image generation techniques, evaluating their strengths and limitations in producing realistic and useful iris images that can be used for both training and testing iris recognition systems and  presentation attack detectors. In this regard, we first survey the various methods that have been used for synthetic iris generation  and specifically  consider generators based on StyleGAN, RaSGAN, CIT-GAN, iWarpGAN, StarGAN, etc. We then  analyze the images generated by these models  for realism, uniqueness, and biometric utility. This comprehensive analysis highlights the pros and cons of various GANs in the context of developing robust iris matchers and presentation attack detectors.
\end{abstract}


\section{Introduction}
\label{sec:intro}
Iris based biometric recognition systems have gained significant attention in recent years with applications in domains such as smartphone access, border control and airport security \cite{nigam2015, gent2023, irisgalaxy8}. However, as this field advances, it brings forth a range of research challenges that require exploration and innovation \cite{ross2019}. Notably, a prominent challenge in biometrics research is the availability of datasets with sufficient size, quality, and diverse intra-class variations. Despite significant strides in biometric technology, the datasets commonly employed for training and evaluating these systems frequently fall short in terms of sample quantity and the inclusion of a broader spectrum of intra-class variabilities. This deficiency hinders the development and dependable evaluation of iris recognition systems. Another critical challenge is to ensure the privacy of individuals whose biometric data is used for research purposes \cite{voigt2017}. Biometric information, being inherently personal and unique, raises concerns about unauthorized access, data breaches, and potential identity theft. Protecting the privacy of individuals while utilizing their biometric data is of utmost importance to foster trust and encourage widespread adoption of biometric systems \cite{voigt2017}. Researchers have been actively working on finding solutions to overcome these challenges. One of such solution is to explore the potential of synthetically generated iris datasets. In fact, synthetic image generation is an active area of research in the field of fingerprint recognition \cite{wyzykowski2023, engelsma2022, zhu2023, suroso2023} and face recognition \cite{osadchy2017, han2018, huang2018, fang2023, taherkhani2023}.

\begin{figure}[t]
 \centering
 \includegraphics[width=.75\linewidth]{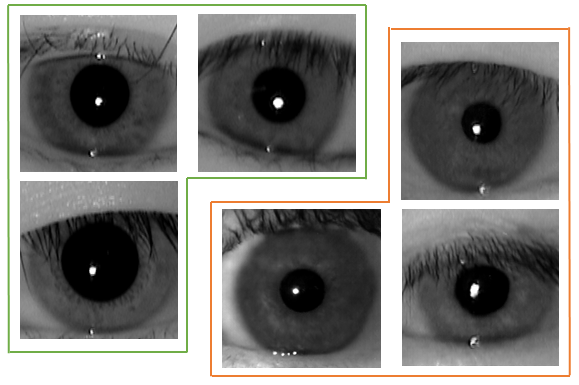}
 \caption{Examples of bonafide irides (green box) and synthetic iris images generated using RaSGAN (red box) \cite{yadav2019}.} 
 \vspace{-4mm}
 \label{fig:motivation}
\end{figure}

Synthetic iris generation refers to the creation of digital iris images that emulate the traits and patterns observed in real iris images \cite{dankar2021, zuo2007, drozdowski2017}. These synthetic samples are designed to possess similar statistical properties as real, providing a valuable resource for research, development, and evaluation in the field of biometrics. Synthetic image generators can be utilized to generate more data with both inter and intra-class variations. This helps overcome the limitations and challenges associated with conventional iris datasets. Traditional iris datasets often suffer from restricted sample sizes, lack diversity, and present concerns regarding privacy and data sharing. In contrast, synthetically generated irides offer a controlled and scalable solution that can mimic the complexity and diversity of real-world iris traits \cite{joshi2022, drozdowski2017}.

By generating synthetic irides, researchers and developers can gain access to larger and more diverse datasets that facilitates comprehensive testing and optimization of iris based biometric algorithms, enhancing their performance, reliability, security and generalization capabilities. Additionally, synthetically generated irides address privacy concerns associated with the utilization of real biometric data. The artificial nature of the generated data ensures that it is not directly linked to any specific individual, mitigating the risks of unauthorized access or misuse of personal information. Consequently, synthetic biometric datasets can be shared and distributed for research and evaluation purposes without compromising individuals' privacy rights. Furthermore, synthetically generated irides play a crucial role in enhancing the training and testing of deep convolutional neural network (CNN) models. Deep CNNs have demonstrated remarkable performance in various biometric tasks but heavily rely on large labeled datasets for effective training. Synthetically generated irides can aid in augmenting the availability of labeled training data by creating synthetic samples with known ground truth annotations. This facilitates the creation of more extensive and diverse training sets, resulting in improved CNN model training and higher accuracy in iris recognition systems. Moreover, synthetically generated irides contribute to the development of robust presentation attack detection (PAD) algorithms. Presentation attacks (PAs), where an adversary attempts to deceive a biometric system using fabricated or altered irides, pose significant security risks. Various presentation attack scenarios can be simulated by generating synthetic irides with diverse variations and attack types (for example, cosmetic contact lens, printed eyes, artificial eyes, etc.). This synthetically generated dataset enables the training and evaluation of PA detection algorithms, enhancing their effectiveness and enabling the development of resilient countermeasures against evolving PA techniques.

Thus, there are various applications of synthetically generated irides \cite{joshi2022, engelsma2022, grosz2022, voigt2017, yadav2021, patrick2023}:

\begin{itemize}
    \item Development and Testing:
    Synthetic iris datasets serve as a valuable resource for algorithm development and testing in the field of biometrics. By generating synthetic iris samples, researchers can evaluate the performance of novel algorithms, compare different techniques, and benchmark their efficacy against a standardized iris dataset. Synthetic data allows for controlled experimentation, enabling researchers to precisely manipulate specific biometric traits, variations, and noise levels to simulate different operational scenarios. Therefore, by using synthetic data, researchers and developers can assess system accuracy, robustness, and vulnerability to various attacks or spoofing attempts. This evaluation process aids in identifying system weaknesses, improving overall system performance, and guiding the development of countermeasures.
    \item Training Data Augmentation:
    Synthetic iris datasets can be utilized to augment training datasets, enhancing the performance of iris recognition systems. By generating additional synthetic samples, researchers can increase the size and diversity of the training set, which helps to improve the generalization capabilities of the algorithms. This approach reduces overfitting, enhances the system's ability to handle intra-class variations, and improves overall recognition accuracy.
    \item Privacy-Preserving:
    Synthetic iris dataset is invaluable for privacy-preserving studies and research involving sensitive biometric information. It allows researchers to conduct studies, simulations, and experiments without the need for real individuals' personal biometric data. Synthetic iris dataset provides a privacy-friendly alternative that ensures data protection while enabling advancements in biometric research and system development.
\end{itemize}

Therefore, synthetic iris datasets offer flexibility and convenience in the development and assessment of iris based recognition systems and attack detection methods. In this context, it is important to understand the different methods employed to generate synthetic iris samples. In this paper, we explore the current state-of-the-art synthetic iris image generation methods and understand the strengths and weaknesses of these methods in terms of the quality, identity uniqueness and utility of the generated images. \footnote{We consider images in the near-infrared (NIR) spectrum for this study.} Thus, this review brings forth specific contributions:

\begin{itemize}
    \item Study the current state-of-the-art methods to generate synthetic irides and explain their pros and cons.
    \item An assessment of synthetic iris images generated by current state-of-the-art methods in terms of quality, uniqueness and utility.
    \item Conduct experiments to analyze deep learning based iris recognition methods and how synthetically generated iris datasets can enhance their performance.
    \item Conduct experiments to analyze current deep learning based presentation attack detection methods and how synthetically generated iris dataset can enhance their performance.
    \item Discuss the future direction to overcome the challenges of current methods to generate enhanced synthetic iris datasets.
\end{itemize}

\begin{figure*}
    \centering
    \includegraphics[width=0.95\textwidth]{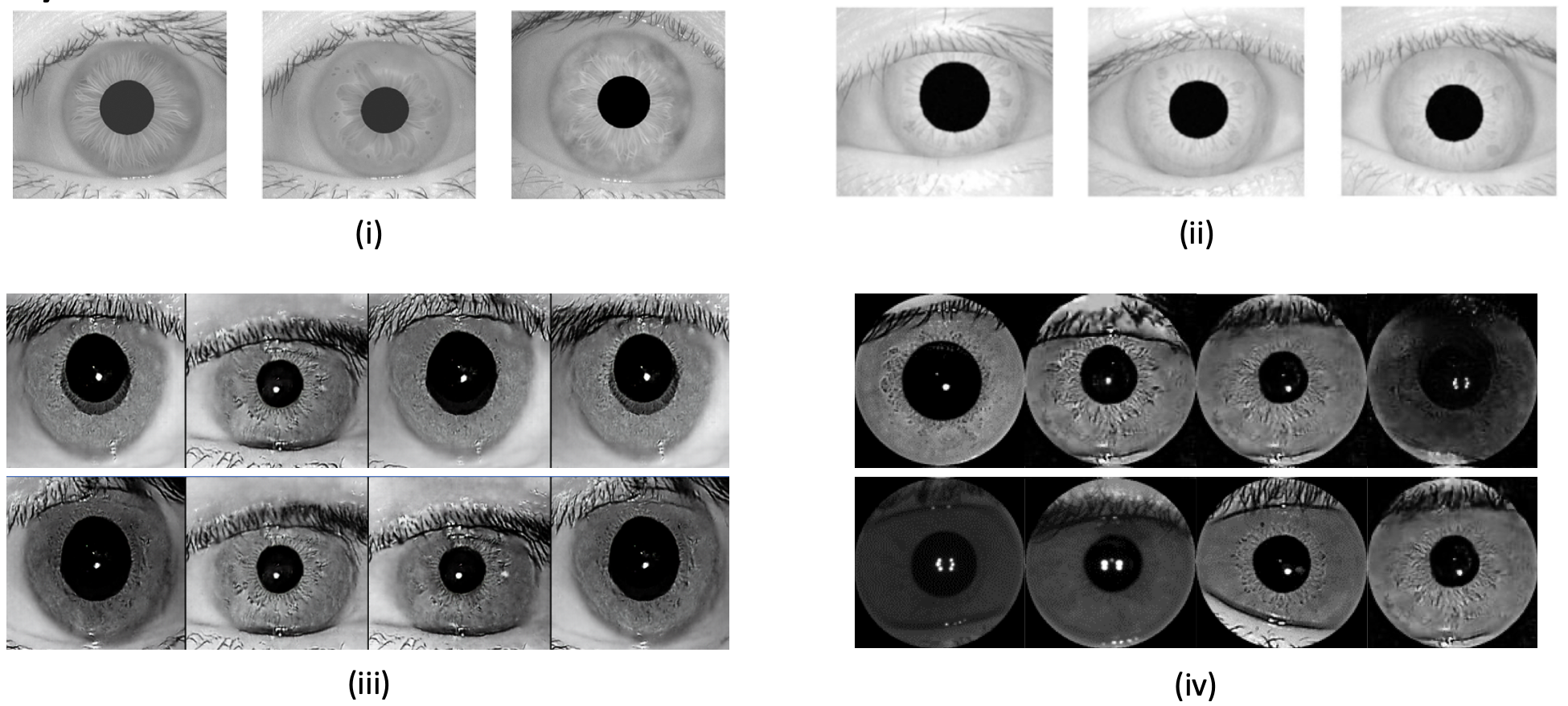}
    \caption{Examples of synthetically generated iris images using different methods. (i) and (ii) use Texture-based non-deep learning methods from \cite{zhou2016} and \cite{shah2006}. On the other hand, (iii) and (iv) use GAN based methods from \cite{yadav2019} and \cite{kohli2017}, respectively. Former models tasked with producing synthetic biometric samples strive to emulate the distribution of real human biometric data, which inherently ties them to the variability and scope of the training datasets. This can lead to challenges in replicating the intricate variations found in real biometric traits, often yielding a synthetic dataset with a narrow range of diversity. Additionally, certain models may fall short in delivering high-quality and lifelike representations. Nonetheless, Generative Adversarial Networks (GANs) have marked a significant advancement in crafting synthetic images that closely resemble real ones. Yet, these images often bear a strong similarity to the training data that comprises actual human images, meaning that they lack distinctiveness in their identities. Moreover, some of these techniques do not incorporate enough intra-class variation, failing to capture the extensive spectrum of natural differences present in authentic biometric data across real-world settings. }
    \label{fig:Recognition}
\end{figure*}

\begin{figure*}
    \centering
    \captionsetup{justification=centering}
    \includegraphics[width=0.95\textwidth]{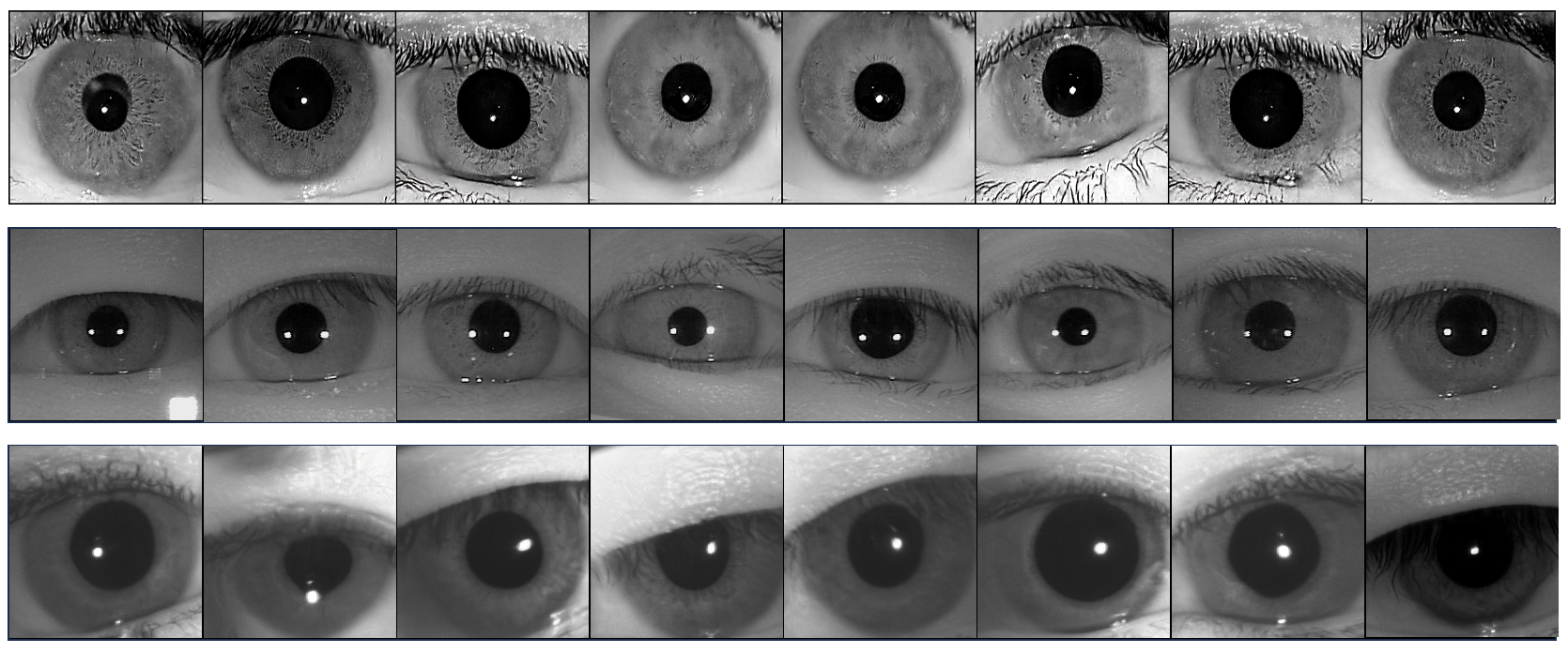}
    \caption{Some more examples of synthetic irides that are generated using different Generative Adversarial Networks (GANs) for partially and fully synthetic iris images.}
    \vspace{-5mm}
    \label{fig:Synth-Samples}
\end{figure*}

\begin{table*}[]
\centering
\label{tab:Dataset}
\caption{Some examples of iris datasets captured in different spectrum using various sensors showcasing the lack of number of samples and also number of subjects in the datasets.}
\scalebox{0.7}{
\begin{tabular}{|c|c|c|c|}
\hline
Dataset Name                                                                & Devices                                                                          & Spectrum               & Number of Images (subjects) \\ 

\hline
CASIA-Iris-Thousand \cite{casia1000}                                                         & Iris scanner (Irisking IKEMB-100)                                                & NIR                    & 20,000 (1000)              \\ 

\hline
CASIA-Iris-Interval \cite{casia1000}                                                         & CASIA close-up iris camera                                               & NIR                    & 2,639 (249)              \\

\hline
CASIA-Iris-Lamp \cite{casia1000}                                                         & OKI IRISPASS-h                                               & NIR                    & 16,212 (411)              \\

\hline
CASIA-Iris-Twins \cite{casia1000}                                                         & OKI IRISPASS-h                                               & NIR                    & 3,183 (200)              \\

\hline
CASIA-Iris-Distance \cite{casia1000}                                                         & CASIA long-range iris camera                                                & NIR                    & 2,567 (142)              \\

\hline
ICE-2005 \cite{ice2005}                                                         & LG2200                                                & NIR                    & 2,953 (132)              \\

\hline
ICE-2006 \cite{ice2006}                                                         & LG2200                                                & NIR                    & 59,558 (240)              \\

\hline
IIITD-CLI \cite{kohli2013}                                                         & Cogent and VistaFA2E single iris sensor                                                & NIR                    & 6,570 (240)              \\

\hline
URIRIS v1 \cite{ubirisv1}                                                         & Nikon E5700                                                & VIS                    & 1,877 (241)              \\

\hline
URIRIS v2 \cite{ubirisv2}                                                         & Canon EOS 5D                                                & VIS                    & 11,102 (261)              \\

\hline
MILES \cite{edwards2012}                                                         & MILES camera                                                & VIS                    & 832 (50)              \\

\hline
MICHE DB \cite{zhang2016}                                                         & iPhone 5, Samsung Galaxy (IV + Tablet II)                                                & VIS                    & 3,732 (184)              \\

\hline
CSIP \cite{santos2015}                                                         & Xperia Arc S, iPhone 4, THL W200, Huawei Ideos X3                                                & VIS                    & 2,004 (100)              \\

\hline
WVU BIOMDATA \cite{crihalmeanu2007}                                                         & Irispass                                                & NIR                    & 3,099 (244)              \\

\hline
IIT-Delhi Iris Dataset \cite{iitd-iris} & \begin{tabular}[c]{@{}c@{}}JIRIS, JPC1000 and\\ digital CMOS camera\end{tabular} & NIR                    & 1,120 (224)                \\ 

\hline
CASIA-BTAS \cite{zhang2016}                                                                 & CASIA Module v2                                                                  & NIR                    & 4,500 (300)                \\ 

\hline
IIITD Multi-spectral Periocular  \cite{sharma2014}                                            & Cogent Iris Scanner                                                                                 & NIR, VIS, Night Vision & 1,240 (62)                \\

\hline
CROSS-EYED  \cite{sequeira2016}                                            & Dual Spectrum Sensor                          & NIR, VIS & 11,520 (240)                \\
\hline
\end{tabular}}
\end{table*}

\section{Iris Recognition}
The foundational technology behind modern automated iris recognition systems can be traced back to the work of John Daugman \cite{daugman2007}, who is credited with the development of the core algorithms that make such systems possible. Daugman's work leverages the distinct patterns found in the human iris for secure and accurate human recognition. There have been significant improvements on his initial work for various security, identification and privacy applications \cite{rankin2009, liu2005, maheswari2008}. Iris Recognition can be divided into 4 stages:

\begin{itemize}
    \item Iris Segmentation: Images in most iris datasets do not only contain the iris itself but also region in the vicinity of the iris such as pupil, sclera, eyelashes etc. So, the first step towards iris recognition is to segment the iris from the captured image to remove unnecessary or extra information. Most of the initial segmentation approaches, including Daugman \cite{daugman1993}, involve identifying the pupil and iris boundaries. In traditional approaches, the occlusions due to eye-lashed and eyelids are minimized by edge detection and curve fitting techniques.
    \item Normalization: Post segmentation, the variations in the size of segmented irides (caused due to distance from sensor, or contraction and dilation of the pupil) are minimized via geometric normalization where annular irides are unwrapped to a fixed-size rectangular image.
    \item Feature Encoding: Iris features are extracted from the geometrically normalized iris image and encoded so that they can be used for matching. Most common techniques for iris feature extraction involve Gabor Filtering \cite{daugman1993}, BSIF \cite{czajka2019}, etc. that help in capturing the unique textural properties of the iris. 
    \item Matching: Once features are encoded, various matching algorithms can be used for iris recognition. Daugman \cite{daugman1993} used Gabor phase information to encode the iris and hamming distance to compare the encoded irides. These have been improved over time to account for variations in image quality, occlusion and noisy data. 
\end{itemize}
For a detailed explanation of iris recognition, we recommend referring to \cite{bowyer2016},
\cite{daugman2007}, \cite{nguyen2017}.

With recent developments in the field of deep-learning, deep networks have found application in all 4 stages of iris recognition. Iris segmentation and feature extraction have particularly benefited from deep learning-based approaches as they often handle noise and complexities in iris datasets better than traditional approaches. Nguyen et al. \cite{nguyen2017} studied the performance of pre-trained convolutional neural networks (CNNs) in the domain of iris recognition. The study reveals that features derived from off-the-shelf CNNs can efficiently capture the complex characteristics of irides. These features are adept at isolating distinguishing visual attributes, which leads to encouraging outcomes in iris recognition performance. While the progress in the field of deep learning has helped improve the reliability of iris recognition systems by improving their performance under various conditions, the lack of iris data with sufficient inter and intra-class variations limits the training and testing of these systems. Therefore, we need to explore the field of generative AI to generate synthetic iris datasets with sufficient inter and intra-class variations to help train and test a robust iris recognition system.

\section{Iris Presentation Attack Detection}
Presentation attack detection (PAD) is an essential component of iris recognition systems. As the reliance on iris recognition systems grows, so does the sophistication of attacks designed to exploit them. Here, we briefly examine the nature of iris presentation attacks, the methodologies developed to detect them, and the challenges faced in enhancing iris PAD systems.

\begin{figure*}[t]
 \centering
 \includegraphics[width=1\linewidth]{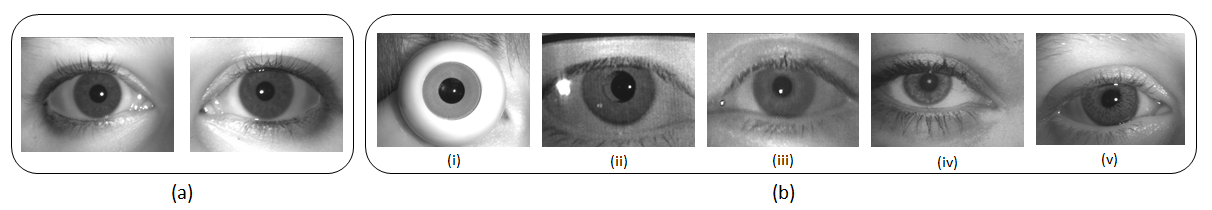}
 \caption{Some examples of real bonafide and PA iris images from MSU-Iris-PA01 \cite{yadav2019}: (a) bonafide samples and (b) presentation attacks: (i) artificial eye, (ii) \& (iii) printed eye, (iv) Kindle display and (v) cosmetic contact lens. \cite{yadav2019, yadav2020}.} 
 \vspace{-4mm}
 \label{fig:Real-PA}
\end{figure*}

Iris presentation attacks (PAs), sometimes known as spoofs, refers to physical artifacts that aims to either impersonate someone or obfuscate one's identity or create a virtual identity (see Figure \ref{fig:Real-PA}). Several types of presentation attacks have been considered in the literature:

\begin{itemize}
    \item Print Attack: One of the simplest forms of iris PA involves the attacker presenting a high-quality photograph of a valid subject's iris to the biometric system. Basic systems might be misled by the photograph's visual fidelity unless they are designed to detect the absence of depth or natural eye movements.
    \item Artificial Eyes: Attackers may employ high-grade artificial (prosthetic or doll) eyes that replicate the iris's texture and three-dimensionality. These artificial eyes seek to deceive scanners that are not sophisticated enough to discern liveness indicators such as the pupil's response to light stimuli.
    \item Cosmetic Contact Lens: A more nuanced approach includes the usage of cosmetic contact lenses that have been artificially created with iris patterns that can either conceal the attacker’s true iris or mimic someone else’s iris. This type of attack attempts to bypass systems that match iris patterns by introducing false textural elements.
    \item Replay-Attack: Playing back a video recording of a bonafide iris to the sensor constitutes another PA. Advanced iris recognition systems counter this by looking for evidence of liveness, like blinking or involuntary pupil contractions.
\end{itemize}

Researchers have proposed different methods to effectively detect different types of PAs. In \cite{Gupta2014, kohli2013} proposed to utilize textual descriptors like GIST, LBP and HOG to detect printed eyes and cosmetic contact lens. Similarly, Raghavendra and Busch \cite{Raghav2015} utilize cepstral features with binary statistical image features (BSIF) to distinguish between bonafide irides and print attacks. Another way to detect print attack is the liveness test that is lacking in printed eyes \cite{czajka2013, kanematsu2007}. Liveness test can also be helpful to detect attacks like artificial eyes. Eye gaze tracking \cite{lee2008} and multi-spectral imaging \cite{chen2012} have good results in detecting printed eyes and artificial eyes. \cite{Hoffman2018, menotti2015} proposed deep network based PA detection methods to detect different types of PAs. To achieve this, Hoffman et al. \cite{Hoffman2018} proposed a deep network that utilizes patch information with a segmentation mask to learn features that can distinguish bonafide from iris PAs. 

While these iris PA detection methods perform well on various datasets, attackers are continuously finding new ways to bypass them, leading to an arms race between security experts and attackers. As a result, the PAD methods need to be constantly updated (re-trained or fine-tuned) and tested against the latest forms of attacks. This calls for PA detection methods that can generalize well over new (or unseen) PAs without the hassle of re-training or fine-tuning. Here, ``Seen PAs" are those which the PAD methods have been exposed to during the training phase. In contrast, ``Unseen PAs" are not included in the training phase, posing a concerning challenge for accurate PA detection. Recent developments in PAD methods have focused on enhancing the ability of systems to generalize, distinguishing bonafide irides from PAs, even when encountering previously unseen PAs. Gupta et al. \cite{gupta2021} proposed a deep network called MVANet, which uses multiple convolutional layers for generalized PA detection. This network not only improves PA detection accuracy, but also addresses the high computational costs typically associated with training deep neural networks by using a simplified base model structure. Evaluations across different databases indicate MVANet's proficiency in generalizing to detect new and unseen PAs. In \cite{sharma2020}, Sharma and Ross proposed D-NetPAD, a PAD method based on DenseNet to generalize over seen and unseen PAs. It has demonstrated a strong ability to generalize across diverse PAs, sensors, and data collections. Their rigorous testing confirms D-NetPAD's robustness in detecting generalized PAs \cite{sharma2020}.

Most of the PAD methods formulate PA detection as a binary-class problem, which demands the availability of a large collection of both bonafide and PA samples to train classifiers. However, obtaining a large number of PA samples can be much more difficult than bonafide iris samples. Further, classifiers are usually trained and tested across similar PAs, but PAs encountered in operational systems can be diverse in nature and may not be available during the training stage. Therefore, we need to explore the generative methods to generate partially synthetic iris images (as identity is not the focus in PA detection) that can help build a balanced iris PA datasets. This will help researchers to better train and test their PAD methods.

\section{Generating Synthetic Irides}
As mentioned earlier, synthetic iris images offer several advantages, including scalability, diversity, and control over the generated data. Some of the methods to generate such images are listed below, categorized on the basis of method used:

\begin{itemize}
    \item Texture Synthesis: This technique has been widely used for generating synthetic iris images. These methods analyze the statistical properties of real iris images and generate new images based on those statistics. Shah and Ross \cite{shah2006} proposed an approach for generating digital renditions of iris images using a two-step technique. In the first stage, they utilized a Markov Random Field model to generate a background texture that accurately represents the global appearance of the iris. In the subsequent stage, various iris features, including radial and concentric furrows, collarette, and crypts, are generated and seamlessly embedded within the texture field. In another example,  Makthal and Ross \cite{makthal2005} introduced an approach for synthetic iris generation using Markov Random Field (MRF) modeling. The proposed method offers a deterministic synthesis procedure, which eliminates the need for sampling a probability distribution and simplifies computational complexity. Additionally, the study highlights the distinctiveness of iris textures compared to other non-stochastic textural patterns. Through clustering experiments, it is demonstrated that the synthetic irides generated using this technique exhibit content similarity to real iris images. In a different approach, Wei et. al. \cite{wei2005} proposed a framework for synthesizing large and realistic iris datasets by utilizing iris patches as fundamental elements to capture the visual primitives of iris texture. Through patch-based sampling, an iris prototype is created, serving as the foundation for generating a set of pseudo irises with intra-class variations. Qualitative and quantitative analyses demonstrate that the synthetic datasets generated by this framework are well-suited for evaluating iris recognition systems.
    \vspace{2mm}

    \item Morphable Models: Morphable models have been utilized for generating synthetic iris images by capturing the shape and appearance variations in a statistical model. These models represent the shape and texture of irises using a low-dimensional parameter space. By manipulating the parameters, synthetic iris images with different characteristics, such as size, shape, and texture, can be generated. Most of the research in this category focuses on generation synthetic iris images with gaze estimation and rendering eye movements. Wood et. al. \cite{wood2016} proposed a 3-D morphable model for the eye region with gaze estimation and re-targeting gaze using a single reference image. Similarly, \cite{banf2009} focuses on achieving photo-realistic rendering of eye movements in 3D facial animation. The model is built upon 3D scans of a face captured from various gaze directions, enabling the capture of realistic motion of the eyeball, eyelid deformation, and the surrounding skin. To represent these deformations, a 3D morphable model is employed.
    \vspace{2mm}

    \item Image Warping: Image warping techniques involve applying geometric transformations to real iris images to generate synthetic images. These transformations can include rotations, translations, scaling, and deformations. Image warping allows for the generation of synthetic iris images with variations in pose, gaze direction, and occlusions. In \cite{cardoso2013iris}, Cardoso et. al. aimed to generate synthetic degraded iris images for evaluation purposes. The method utilizes various degradation factors such as blur, noise, occlusion, and contrast changes to simulate realistic and challenging iris image conditions. The degradation factors are carefully controlled to achieve a realistic representation of degraded iris images commonly encountered in real-world scenarios. In \cite{cui2004}, a method combining principal component analysis (PCA) and super-resolution techniques is proposed. The study begins by introducing the iris recognition algorithm based on PCA, followed by the presentation of the iris image synthesis method. The proposed synthesis method involves the construction of coarse iris images using predetermined coefficients. Subsequently, super-resolution techniques are applied to enhance the quality of the synthesized iris images. By manipulating the coefficients, it becomes possible to generate a wide range of iris images belonging to specific classes.
    \vspace{2mm}

    \item Generative Adversarial Networks (GANs): GANs have gained significant attention for generating realistic and diverse synthetic iris images. In a GAN framework, a generator network learns to generate synthetic iris images, while a discriminator network distinguishes between real and synthetic images. The two networks are trained in an adversarial manner, resulting in improved image quality over time. GANs can generate iris images with realistic features, including iris texture, color, and overall appearance. Minaee and Abdolrashidi \cite{minaee2018} proposed a framework that utilizes a generative adversarial network (GAN) to generate synthetic iris images sampled from a learned prior distribution. The framework is applied to two widely used iris datasets, and the generated images demonstrate a high level of realism, closely resembling the distribution of images within the original datasets. Similarly, Kohli et. al. \cite{kohli2017} proposed iDCGAN (iris Deep Convolutional Generative Adversarial Network), a novel framework that leverages deep convolutional generative adversarial networks and iris quality metrics to generate synthetic iris images that closely resemble real iris images. Bamoriya et. al. \cite{bamoriya2022} proposed an novel approach, called Deep Synthetic Biometric GAN (DSB-GAN), for generating realistic synthetic biometrics that can serve as large training datasets for deep learning networks, enhancing their robustness against adversarial attacks. In \cite{bhuiyan2024}, Bhuiyan et. al. proposed to advance post-mortem iris recognition by presenting a unique iris synthesis model based on StyleGAN that aims to study and manipulate the latent space of StyleGAN to generate post-mortem iris images, maintaining consistency within the same identity while also creating diverse identities.
    \vspace{2mm}
\end{itemize}

Currently, GAN-based methods to generate synthetic biometrics have been proven to be far superior to capture the intricate details of various biometric cues. Therefore, in the remaining of the paper we will mainly focus on these methods and iris images generated by these methods for our study and analysis.

\subsection{Partially Synthetic Irides}
\begin{figure*}
    \centering
    \captionsetup{justification=centering}
    \includegraphics[width=0.9\textwidth]{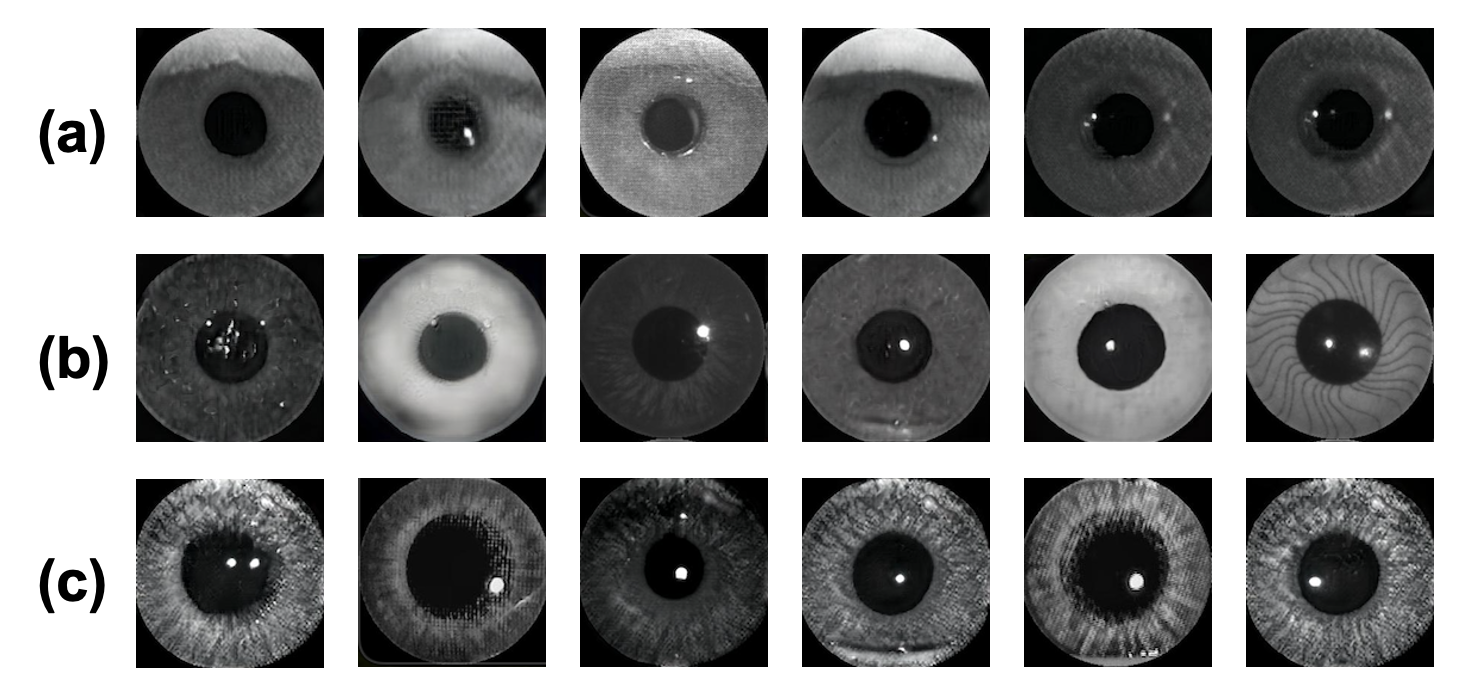}
    \caption{Some examples of partially-synthetic iris PAs (a: Printed eyes, b: artificial eyes and c: cosmetic contact lens) generated using CIT-GAN \cite{yadav2023}.}
    \vspace{-5mm}
    \label{fig:PA-generated}
\end{figure*}

 Partially-Synthetic data refer to synthetic samples that contain artificial components mixed with real biometric data. In this approach, certain aspects or attributes of the biometric data are synthetically generated, while other parts are derived from real individuals (see Figure \ref{fig:PA-generated}). The goal of partially-synthetic data is to introduce controlled variations or augmentations to the real data, thereby increasing the diversity and robustness of the dataset. This can be particularly useful in scenarios where the real data is limited, imbalanced, or lacks specific variations. For example, in iris presentation attack (PA) detection where the detection methods aim to detect PA attacks (such as printed eyes, cosmetic contact lens, etc.), limited PA data is available to train the detection methods. This can limit the methods' development and testing as well. Also, with the improvement in technology more advance PA attacks are present in the real world (such as good quality textured contact lens, replay attack using high definition screens, etc.) and the current detection methods are not generalized enough to detect these new and unseen attacks. As mentioned earlier, Kohli et al. \cite{kohli2017} proposed iDCGAN that utilizes a deep convolutional generative adversarial network to generate synthetic iris images that are realistic looking and closely resemble real iris images. This framework aims to explore the impact of the synthetically generated iris images when used as presentation attacks on iris recognition systems. In \cite{bamoriya2022}, Bamoriya et al. proposed a novel approach, DSB-GAN, which is built upon combination of convolutional autoencoder (CAE) and DCGAN and the evaluation of DSB-GAN is conducted on three biometric modalities: fingerprint, iris, and palmprint. One of the notable advantages of DSB-GAN is its efficiency due to a low number of trainable parameters compared to existing state-of-the-art methods. Yadav et al. \cite{yadav2019,yadav2020} leverages RaSGAN to generate high-quality partially synthetic iris images in NIR spectrum and evaluates the effectiveness and usefulness of these images as both bonafide and presentation attack. They also proposed a novel one-class presentation attack detection method known as RD-PAD for unseen presentation attack detection, addressing the challenge of generalizability in PAD algorithms. Zou et al. \cite{zou2018} proposed 4DCycle-GAN that is designed to enhance the database of iris PA images by generating synthetic iris images with cosmetic contact lenses. Building upon the Cycle-GAN framework, the 4DCycle-GAN algorithm stands out by adding two discriminators to the existing model to increase the diversity of the generated images. These additional discriminators are engineered to favor the images from the generators rather than those from real-life captures. This approach reduces the bias towards generating repetitive textures of contact lenses, which typically make up a significant portion of the training data. In \cite{yadav2021}, Yadav and Ross proposed to generate bonafide as well as different types of presentation attacks in NIR spectrum using a novel image translative GAN, known as CIT-GAN. The proposed architecture translates the style of one domain to another using a styling network to generate realistic, high resolution iris images.

\subsection{Fully Synthetic Irides}
\begin{figure*}
    \centering
    \captionsetup{justification=centering}
    \includegraphics[width=0.9\textwidth]{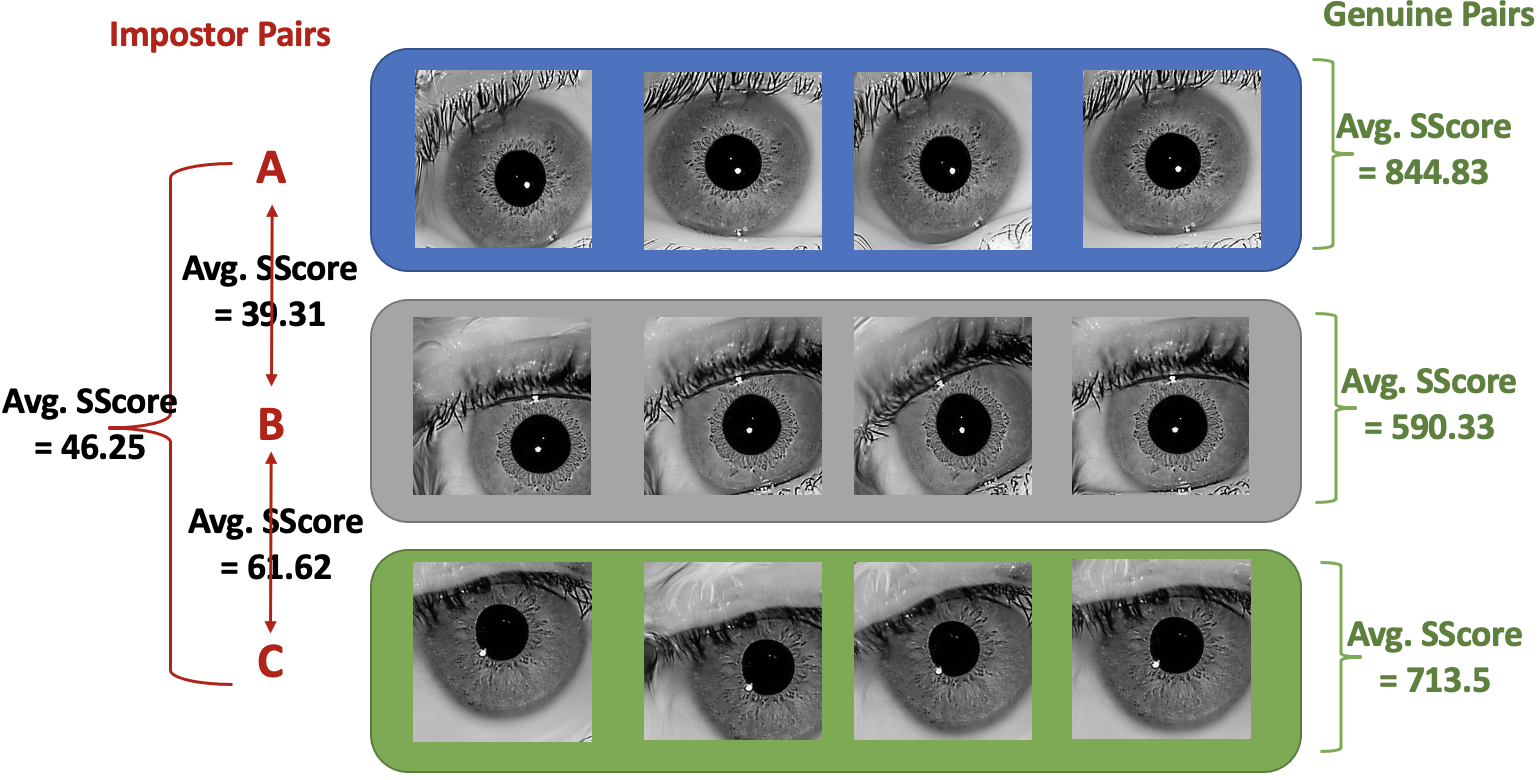}
    \caption{Some examples of fully-synthetic irides generated using iWarpGAN with inter and intra-class variations \cite{yadav2023}. SScore here refers to simiarity score between two iris images using VeriEye.}
    \vspace{-5mm}
    \label{fig:generated}
\end{figure*}

Fully-synthetic biometric data refer to entirely artificial biometric samples that do not correspond to any real individuals in the training data. These synthetic samples are created using mathematical models, statistical distributions, or generative algorithms to simulate the characteristics of real biometric data (see Figure \ref{fig:generated}). Some of the texture based methods focuses on generating new iris identities (fully-synthetic) that are unique from training samples. These methods aim to generate new iris images with both inter and intra-class variations to help mitigate the issue of small training data size by increasing the size of the dataset. This can help in the development of recognition systems and their testing. Also, by generating fully-synthetic identities that do not correspond with real identities, we can solve the privacy concerns associated with using a real person's biometric data. Wang et al. \cite{wang2022} proposed a novel algorithm for generating diverse iris images, enhancing both the variety and the number of images available for analysis. The technique employs contrastive learning to separate features tied to identity (like iris texture and eye orientation) from those that change with conditions (such as pupil size and iris exposure). This separation allows for precise identity representation in synthetic images. The algorithm uniquely processes iris topology and texture through a dual-channel input system, enabling the generation of varied iris images that retain specific texture details. Yadav and Ross \cite{yadav2023} proposed iWarpGAN that aims to disentangle identity and stylistic elements within iris images. It achieves this through two distinct pathways: one that transforms identity features from real irides to create new iris identities, and another that captures style from a reference image to infuse it into the output. By merging these modified identity and style elements, iWarpGAN can produce iris images with a wide range of inter and intra-class variations. Limited work has been done in this category to generate irides with identity that do not match with any identity in the training data. This is an important and upcoming topic that needs more focus.

\section{Generating Adversarial Networks (GANs)}
A generative adversarial network (GAN) consists of two networks: a generator and a discriminator, which are trained simultaneously through adversarial learning. The generator network aims to generate data, such as images, audio, or text, that is indistinguishable from real data, while the discriminator aims to differentiate between real and generated data. In this study, we have utilized five different types of GANs that has been shown in the literature to generate realistic looking iris images:

\begin{itemize}
    \item \textbf{Relativistic Average GAN (RaSGAN)}: RaSGAN \cite{jolicoeur2018} aims to overcome the shortcomings of traditional GANs by introducing a relativistic discriminator. In standard GANs, the discriminator's objective is to distinguish between real and fake samples, while the generator's objective is to generate samples that are indistinguishable from real ones. However, in RaSGAN, both the generator and discriminator are trained to consider the relative likelihood that a real sample is more realistic than a synthetic one and vice-versa. The aim is to provide the discriminator with feedback not only on from the real and synthetic samples but also on how realistic each sample is with respect to each other. Thus, improving the stability and convergence properties of GAN by providing more informative gradients to both the generator and discriminator. RaSGANs have been shown to produce realistic synthetic images across various domains, including images \cite{jolicoeur2018, yadav2019}, text \cite{nam2020}, and audio \cite{zhang2020, nakatsuka2021}.
    \vspace{2mm}
    
    \item \textbf{StarGAN-v2}: Choi et al. \cite{choi2020} proposed a multi-domain image translative GAN knowns as StarGAN-v2 that aims to generate realistic looking images in multiple domains. The domains here refers to styles such as hair color, facial expression, etc. The generator in StarGAN-v2 takes a source image and a reference style code generated by mapping network as inputs and translate the source image into a synthetic image that is from domain exhibiting style properties same as reference style code. This method showed improvement for multi-domain synthetic image generation when compared with StyleGAN \cite{karras2019} and StarGAN \cite{choi2018}.
    \vspace{2mm}
    
    \item \textbf{Cyclic Image Translative GAN (CIT-GAN)}: Yadav and Ross \cite{yadav2020} proposed CIT-GAN to generate different types of iris PAs as well as bona-fide images. In this method, the generator aims to translate given input image into an image from reference domain. This translation is facilitated by a Style Network that takes reference image as input and actively learns to generate style code that clearly indicates the style properties of reference domain. In comparison to StarGAN-v2 that uses mapping network to generate style code from a random noise vector, this method is able to learn the intra-class variations that are present in the multiple domains. Therefore, improving the diversity in the generated images for each domain. The "cyclic" term here refers to cycle-consistency loss that is utilized to ensure that the translated image can be translated back to the original source domain. This helps ensure that while translating the image from source domain to reference, underlying characteristic of the input image such as iris-pupil shape and size are not changed.
    
    \item \textbf{StyleGAN-3}: Karras et al. first proposed StyleGAN \cite{karras2019} for good quality synthetic image generation with some improvements in StyleGAN-2 \cite{karras2020} to handle multi-domain image generation, but these approaches lack in terms of diversity in images being generated and generalization capability of the network. In \cite{karras2021}, Karras et al. proposed StyleGAN-3 that overcome these shortcomings by introducing an adaptive discriminator augmentation (ADA), which aims at improving the generalization capability of the discriminator network. ADA dynamically adjusts the strength of data augmentation during discriminator training, effectively making the discriminator more robust to diverse variations in the training data.
    \vspace{2mm}
    
    \item \textbf{iWarpGAN}: In \cite{yadav2023}, Yadav and Ross proposed iWarpGAN that aims to generate fully  synthetic iris images i.e., the identity in the iris images generated by this method doesn't resemble any identity seen during training. This is achieved by disentangling identity and style using two transformation pathways: (1) Style transformation pathway: aims to generate iris images with styles that are extracted from a reference image, without altering the identity (2) Identity transformation pathway: aims to transform identity of input image in the latent space to generate identities that is different from input image. This method can generate iris images with identities unique from training data, generate multiple images per newly generated identity and scalable to hundred-thousand identities.
    
\end{itemize}

    \begin{figure*}
        \centering
        \includegraphics[width=0.98\linewidth]{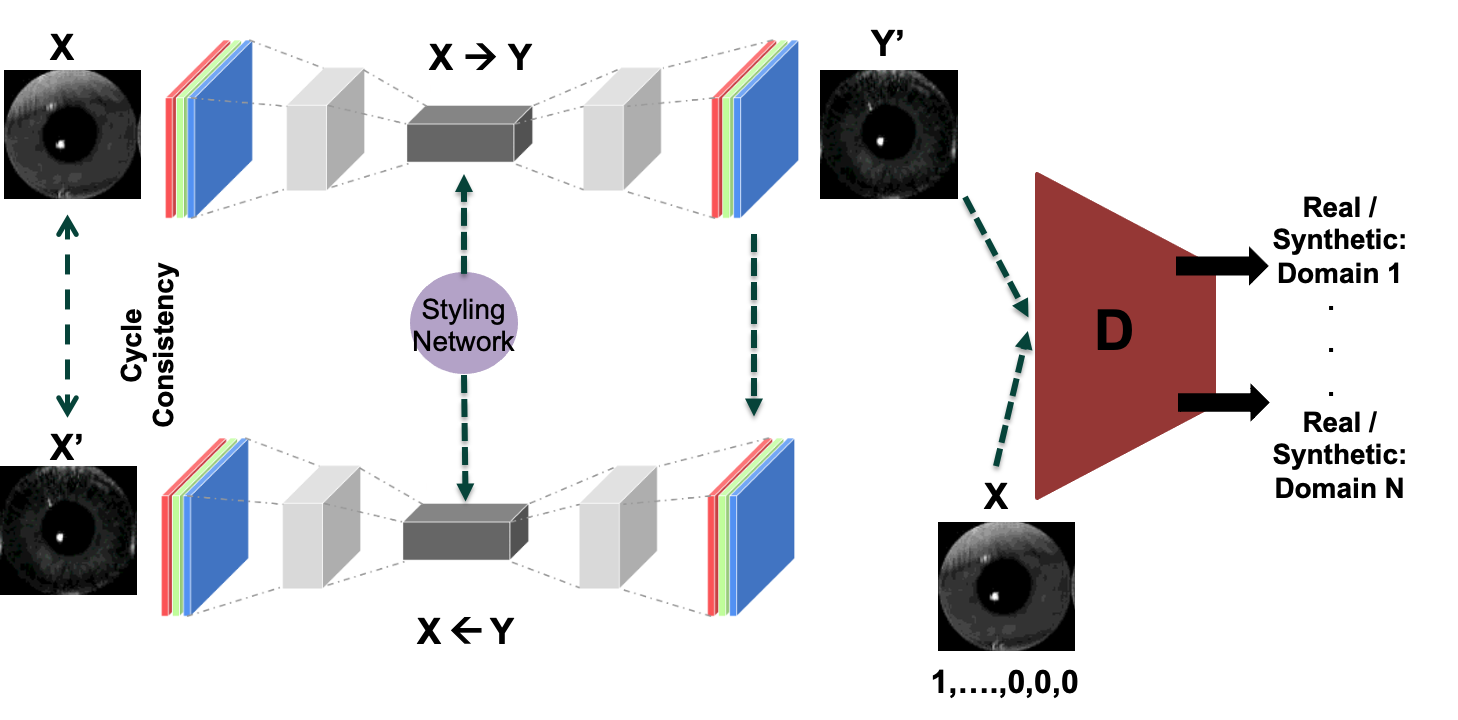}
        \caption{Yadav and Ross \cite{yadav2020} proposed CIT-GAN where the generator aims to translate given input image into an image from reference domain. This translation is facilitated by a Style Network that takes reference image as input and actively learns to generate style code that clearly indicates the style properties of reference domain}
        \label{fig:CIT-Arch}
    \end{figure*}
    \vspace{2mm}

    \begin{figure*}
        \centering
        \includegraphics[width=0.98\linewidth]{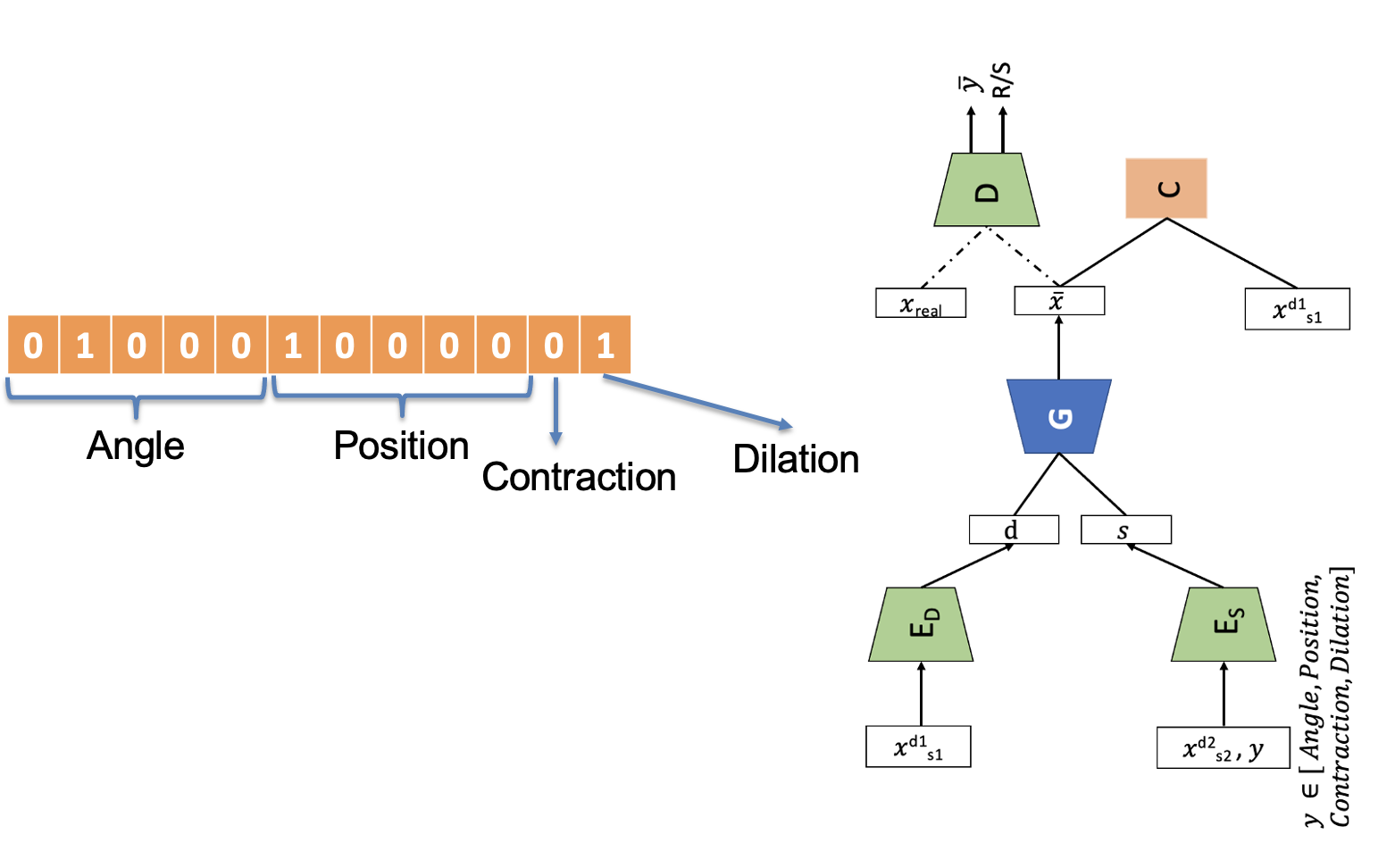}
        \caption{Yadav and Ross \cite{yadav2023} proposed iWarpGAN that aims to generate fully  synthetic iris images i.e., the identity in the iris images generated by this method doesn't resemble any identity seen during training. This is achieved by disentangling identity and style using two transformation pathways: (1) Style transformation pathway: aims to generate iris images with styles that are extracted from a reference image, without altering the identity (2) Identity transformation pathway: aims to transform identity of input image in the latent space to generate identities that is different from input image.}
        \label{fig:iWarpGAN-Arch}
    \end{figure*}
    \vspace{2mm}

\section{Comparative Analysis}
In this section, we discuss the different experiments that were conducted to evaluate the capability of different GAN methods for generating fully and partially synthetic iris images. For this, we test the realism of generated images with respect to real iris data using different methods. Further, uniqueness of the generated images in terms of identity and utility of the generated images for PA detection and iris recognition is evaluated in this section. The different GAN methods we studied in this research are: RaSGAN \cite{jolicoeur2018, yadav2019}, CIT-GAN \cite{yadav2020}, StarGAN-v2, StyleGAN-3 \cite{zhu2023} and iWarpGAN \cite{yadav2023}.

\subsection{Datasets Used}
In this paper, we conducted our experiments and analysis using three iris datasets:

\begin{itemize}
    \item CASIA-Iris-Thousand \cite{casia1000}: Developed by the Chinese Academy of Sciences Institute of Automation, the CASIA-Iris-Thousand dataset is a popular resource for studying iris patterns and for advancing iris recognition technologies. This dataset comprises 20,000 iris images from 1,000 participants, accounting for 2,000 distinct identities when considering images of both the left and right eyes. These images are captured at a resolution of 640x480 pixels. The dataset has been partitioned into training and testing subsets, with a distribution of 70\% for training (1,400 identities) and 30\% for testing (600 identities).
    \item CASIA Cross Sensor Iris Dataset (CSIR) \cite{xiao2013}: The training portion of the CASIA-CSIR dataset, provided by the Chinese Academy of Sciences Institute of Automation, was employed in our study. It includes a total of 7,964 iris images from 100 individuals, representing 200 unique identities when both eyes are considered. Similar to the first dataset, a 70-30 split based on unique identities was used to divide the images into training (5,411 images) and testing sets (2,553 images), intended for the training and evaluation of deep learning models for iris recognition.
    \item IITD \cite{iitd-iris}: Originating from the Indian Institute of Technology, Delhi, the IITD-iris dataset was collected in an indoor setting and consists of 1,120 iris images from 224 subjects. The images were captured using JIRIS JPC1000 and digital CMOS cameras, each with a resolution of 320x240 pixels. In line with the previous datasets, this one also utilizes a 70-30 split based on unique identities for its training (314 identities) and testing sets (134 identities).
\end{itemize}

For the preparation of the data, we processed and resized all the iris images to 256x256 pixels, centering on the iris region as determined by the iris and pupil coordinates from VeriEye.\footnote{\url{www.neurotechnology.com/verieye.html}}
\vspace{1mm}

\subsubsection{Iris PA Datasets}
Our exploration into synthetic images for iris PA detection involves leveraging five distinct iris PA datasets. These datasets included Casia-iris-fake \cite{sun2014}, Berc-iris-fake \cite{lee2007}, NDCLD15 \cite{doyle2015}, LivDet2017 \cite{yambay2015}, and MSU-IrisPA-01 \cite{yadav2019}, each comprising authentic iris images alongside various categories of PAs such as cosmetic contacts, printed iris images, artificial eyes, and display-based attacks. As mentioned earlier, we processed and resized the images to 256x256 pixels, centering on the iris region as determined by the iris and pupil coordinates from VeriEye. Images that VeriEye failed to process correctly were excluded from the study, with our focus being primarily on the aspect of image synthesis. The resulting processed dataset for analysis contains 24,409 genuine iris images, 6,824 images with cosmetic contact lenses, 680 artificial eye representations, and 13,293 printed iris images. The train and test division on this dataset is explained later in this section.

\subsection{Quality of Generated Images}
In order to evaluate the realism and quality of the generated iris images, different GAN methods- RaSGAN, CIT-GAN, StarGAN-v2, StyleGAN-3 and iWarpGAN- are trained using real irides from CASIA-Iris-Thousand, CASIA Cross Sensor Iris and IITD-iris dataset, separately. Using the trained networks, we generate three sets of 20,000 synthetic bonafide images (from each dataset) for each of the GANs mentioned above. We then evaluate the realism of the generated images and the quality of the iris using three different methods: (1) Fréchet Inception Score \cite{salimans2016}, (2) VeriEye Rejection Rate and (3) ISO-ISO/IEC 29794-6 Standard Quality Metrics \cite{iso-quality}.

\begin{figure*}
\centering
\subcaptionbox{FID score distribution from generated iris images when GANs are trained using CASIA-Iris-Thousand dataset.}{\includegraphics[width=0.85\textwidth]{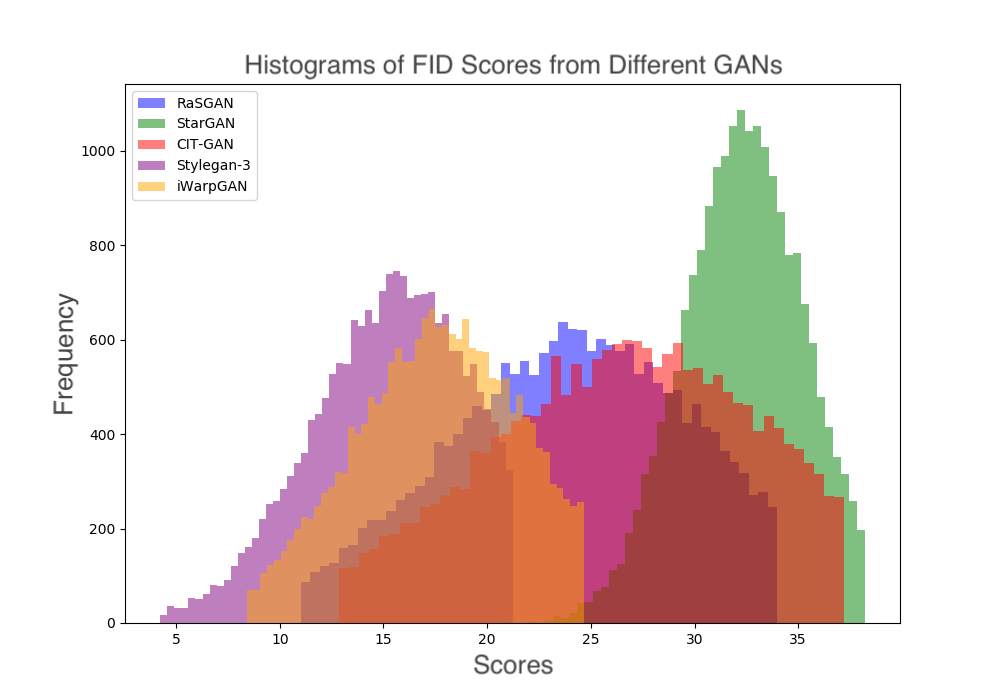}}\quad
\subcaptionbox{FID score distribution from generated iris images when GANs are trained using CASIA-CSIR dataset.}{\includegraphics[width=0.85\textwidth]{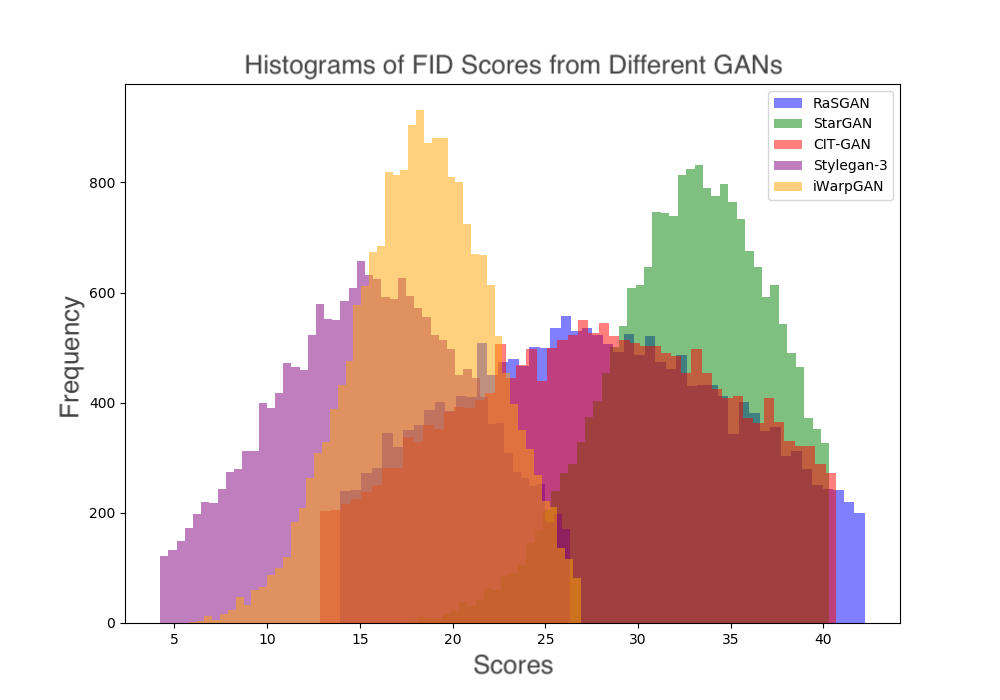}}
\caption{Histograms showing the realism scores of real iris images (i.e., FID scores) from three different datasets and the synthetically generated iris images (continued). Lower the FID score, the more realistic the generated iris images are.}
\end{figure*}

\begin{figure*}
\centering
\ContinuedFloat 
\subcaptionbox{FID score distribution from generated iris images when GANs are trained using IITD-iris dataset.}{\includegraphics[width=0.85\textwidth]{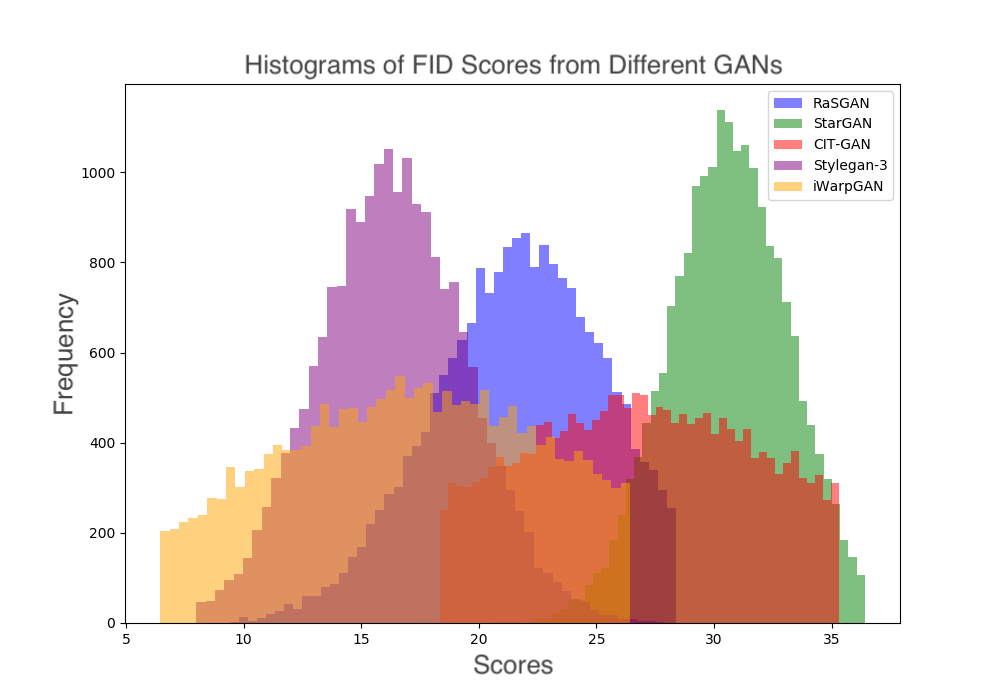}}
\caption{Histograms showing the realism scores of real iris images (i.e., FID scores) from three different datasets and the synthetically generated iris images (continued). Lower the FID score, the more realistic the generated iris images are.}
\label{fig:Image-Quality-FID}
\end{figure*}

\subsubsection{Fréchet Inception Distance Score}
The Fréchet Inception Distance (FID) Score is a metric used to assess the quality of synthetically generated images by comparing their distribution to that of real images, resulting in a score based on the differences. The objective is to minimize this score, as a lower FID score suggests greater resemblance between the synthetic and real datasets. It has been noted that FID scores can span a broad range, with extremely high scores in the 400-600 range indicating significant deviation from the real data distribution and, consequently, poor synthetic image quality \cite{salimans2016}.

In our specific analysis of the quality of synthetically generated iris images produced by different GANs used in this study, we obtained an average FID score of 24.33 and for RaSGAN and StarGAN-v2. On the other hand, a score of 31.82, 26.90, 15.72 and 17.62 were obtained for CIT-GAN, StyleGAN-3 and iWarpGAN, respectively. As mentioned earlier, lower the FID score, more realistic the generated images are with respect to real images. Therefore, we conclude that StyleGAN-3 and iWarpGAN generate the most realistic iris images. The distribution of these FID scores has been shown in Figure \ref{fig:Image-Quality-FID}.

\begin{figure*}
\centering
\subcaptionbox{CASIA-Iris-Thousand dataset versus synthetic iris images from various GANs.}{\includegraphics[width=0.95\textwidth]{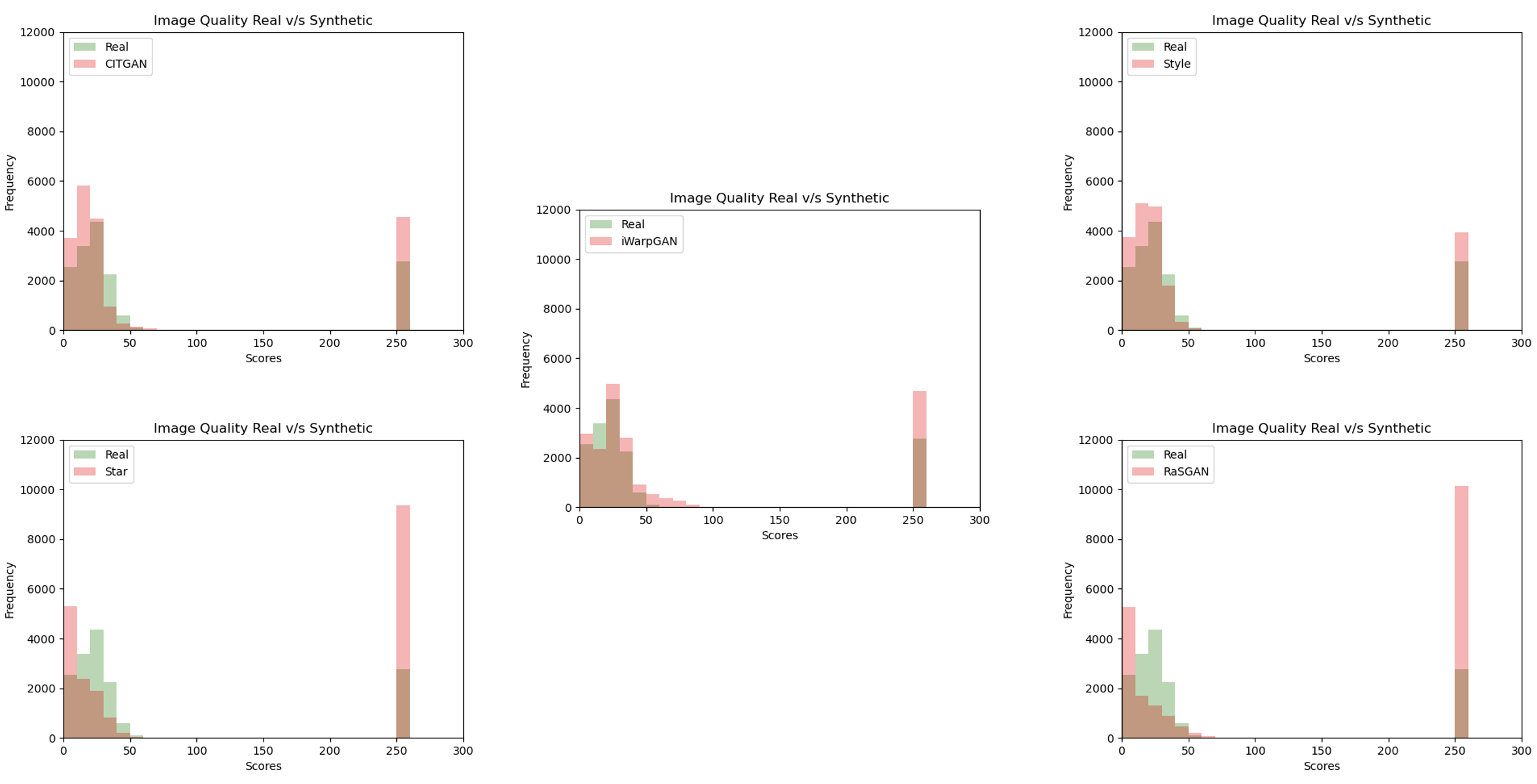}}\quad
\subcaptionbox{CASIA-CSIR dataset versus synthetic iris images from various GANs.}{\includegraphics[width=0.95\textwidth]{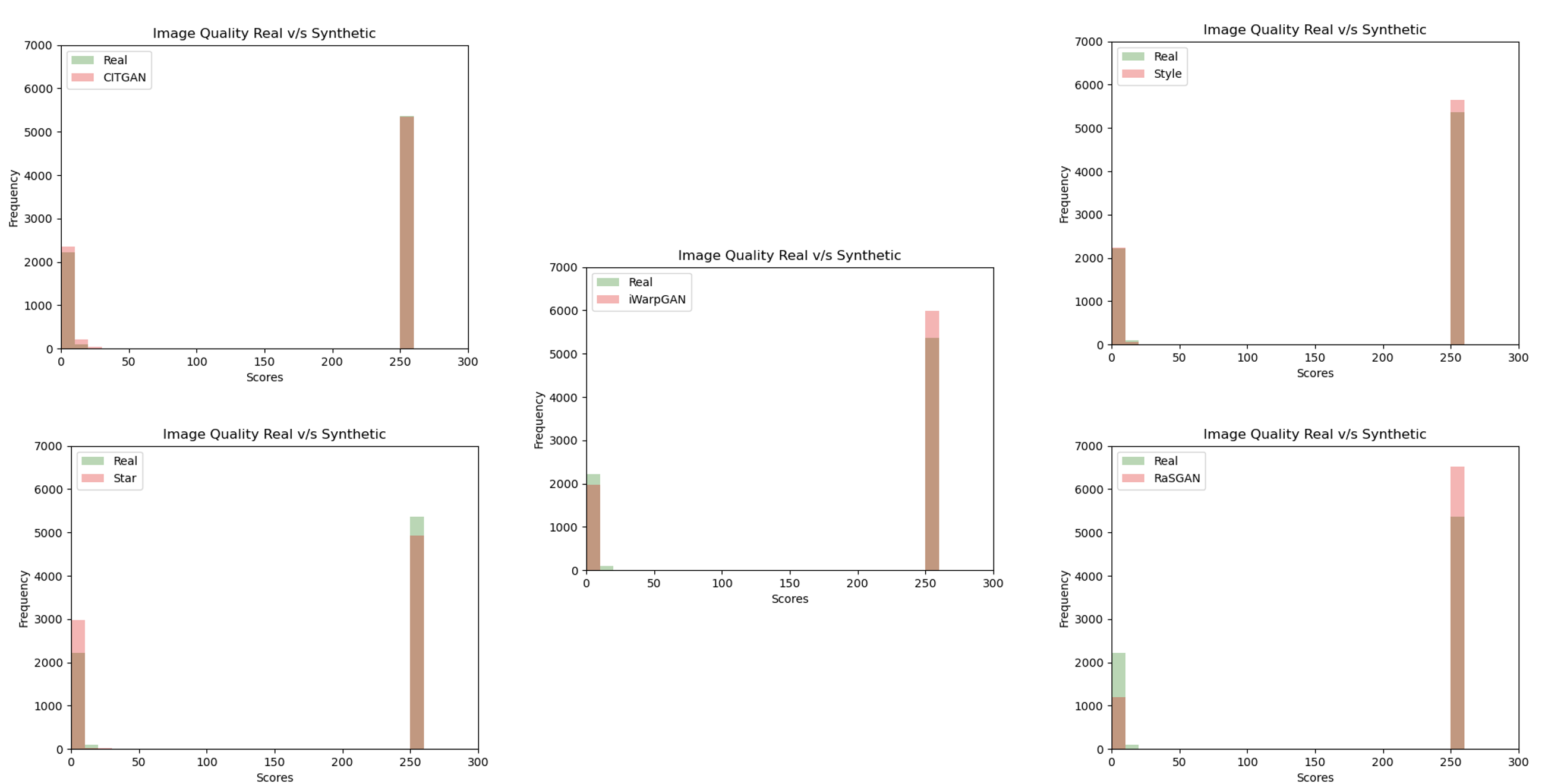}}
\caption{Histograms depicting the quality of real irides alongside the quality of synthetic irides. These evaluations are in accordance with the ISO/IEC 29794-6 Standard Quality Metrics, with the quality scale set between 0 and 100, where a higher score denotes superior quality. Iris images that were not successfully assessed by this standard were assigned a score of 255. As seen from the figure, score distribution for real images are closely resembled by images generated using iWarpGAN, StyleGAN-3, followed by CIT-GAN. However, same can not be said for RaSGAN and StarGAN-v3.}
\label{fig:Exp1}
\end{figure*}

\begin{figure*}
\centering
\ContinuedFloat 
\subcaptionbox{IITD-iris dataset versus synthetic iris images from various GANs.}{\includegraphics[width=0.95\textwidth]{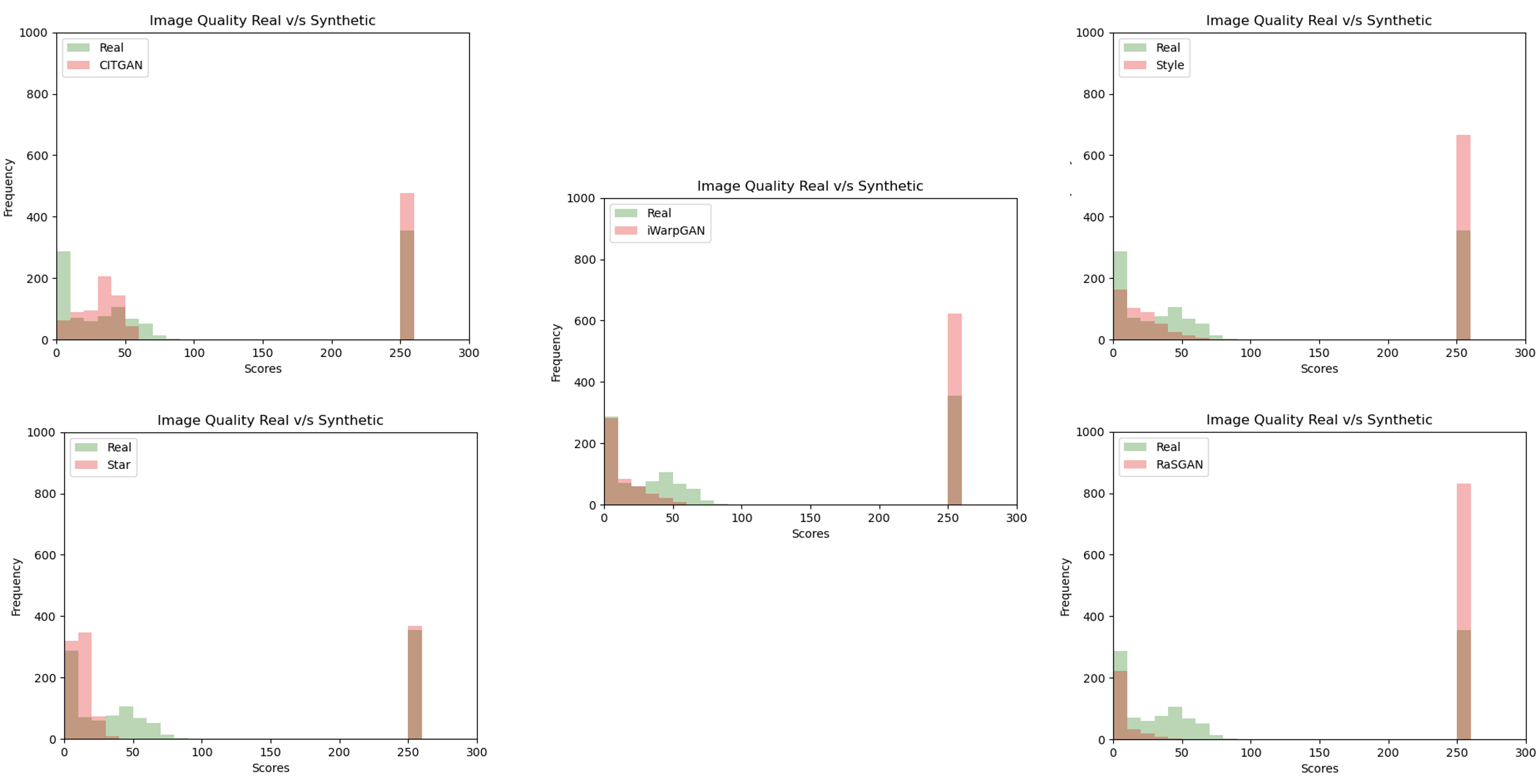}}
\caption{Histograms depicting the quality of real irides alongside the quality of synthetic irides. These evaluations are in accordance with the ISO/IEC 29794-6 Standard Quality Metrics, with the quality scale set between 0 and 100, where a higher score denotes superior quality. Iris images that were not successfully assessed by this standard were assigned a score of 255. As seen from the figure, score distribution for real images are closely resembled by images generated using iWarpGAN, StyleGAN-3, followed by CIT-GAN. However, same can not be said for RaSGAN and StarGAN-v3.}
\label{fig:ISO-Exp}
\end{figure*}

\subsubsection{VeriEye Rejection Rate}
For this experiment, we followed the protocol mentioned in \cite{yadav2023} to evaluate the effectiveness of the various generators by computing the rejection \footnote{Proportion of images that could not be processed.} rate of the images when analyzed by VeriEye, a commercial iris matcher. In the first set of comparisons using the IITD-Iris-Dataset, which comprises 1,120 real iris images, only 0.18\% were rejected by VeriEye. In contrast, 1,120 synthetic images produced by RaSGAN and StarGAN-v2 had a rejection rate of 4.55\% and 4.64\%, respectively. However, images generated by CIT-GAN, iWarpGAN and StyleGAN-3 demonstrated significantly lower rejection rates at 2.85\%, 0.73\% and 1.07\%, respectively.

The CASIA-CS Iris dataset contains 7,964 real iris images with a rejection rate of 2.81\%. The rejection rates for 7,964 synthetic images were notably higher; images generated by RaSGAN and StarGAN-v2 were rejected at a rate of 2.06\% and 2.65\%, respectively. Meanwhile, CIT-GAN, iWarpGAN and StyleGAN-3 produced images with rejection rates closer to the real images, at 2.71\%, 2.74\% and 2.52\% respectively.

Lastly, the CASIA-Iris-Thousand Dataset, which included 20,000 real iris images, saw a very low rejection rate of 0.06\%. Synthetic images from this dataset indicated that RaSGAN and StarGAN-v2 had the highest rejection rate at 0.34\% and 0.27\%, respectively. Synthetic images from CIT-GAN, iWarpGAN and StyleGAN-3 showed improvement with rejection rates of 0.24\%, 0.18\% and 0.16\%, respectively.

\subsubsection{ISO/IEC 29794-6 Standard Quality Metrics}
As described in \cite{yadav2023}, the fidelity of synthetically produced iris images is assessed using the ISO/IEC 29794-6 Standard Quality Metrics \cite{iso-quality}. This assessment was applied to images generated by different GANs utilized in this study. The ISO standard employs a set of criteria to evaluate the quality of an iris image, which includes the usable iris area, the contrast between the iris and sclera, image sharpness, the contrast between the iris and pupil, pupil shape, and more, culminating in a comprehensive quality score. This score is on a scale from 0 to 100, where 0 indicates the lowest image quality and 100 the highest. Images that fail to be evaluated by this ISO metric, typically due to substandard quality or errors in segmentation, are assigned a score of 255.

As shown in Figure \ref{fig:ISO-Exp}, the quality scores for 20,000 synthetic iris images obtained using iWarpGAN, CIT-GAN and StyleGAN-3 are on par with those of real iris images. Conversely, a noticeable number of images generated by RaSGAN were assigned the score of 255, reflecting their inferior quality. Additionally, a comparison across the three datasets showed that the CASIA-CSIR dataset contained a higher proportion of images with the lowest score of 255, in contrast to the IITD-iris and CASIA-Iris-Thousand datasets.

\subsection{Uniqueness of Synthetically Generated Irides}
This experiment examines the uniqueness of the iris images generated synthetically using different GANs, specifically assessing the ability of these methods to create distinct identities that exhibit some intra-class variations. For this, RaSGAN, CIT-GAN, StarGAN-v2, StyleGAN-3 and iWarpGAN- are trained using real irides from the train set of CASIA-Iris-Thousand, CASIA Cross Sensor Iris and IITD-iris dataset, separately.
\vspace{2mm}

Unique-Experiment-1: This experiment is centered on exploring the uniqueness of the synthetically generated iris datasets, which were generated using various GAN techniques, with respect to the training examples. To achieve this, the analysis involve comparing the impostor and genuine distribution for real irides that were part of the training set with that of the synthetically generated irides. Here, VeriEye matcher is utilized to obtain the similarity score between two iris images, with the score spanning from 0 to 1557, where a higher score indicates a more accurate match.
\vspace{2mm}

Unique-Experiment-2: This experiment aims to investigate the uniqueness and intra-class variability present in the synthetically generated iris dataset. This involves an analysis of both genuine and imposter distributions within the generated dataset and their comparison with the distributions from actual iris datasets. As previously noted, this investigation is conducted across a range of uniquely generated identities to assess their uniqueness. The VeriEye matcher is used in this experiment for assessing the similarity score between pairs of iris images.
\vspace{2mm}

\begin{figure*}
    \centering
    \includegraphics[width=0.85\textwidth]{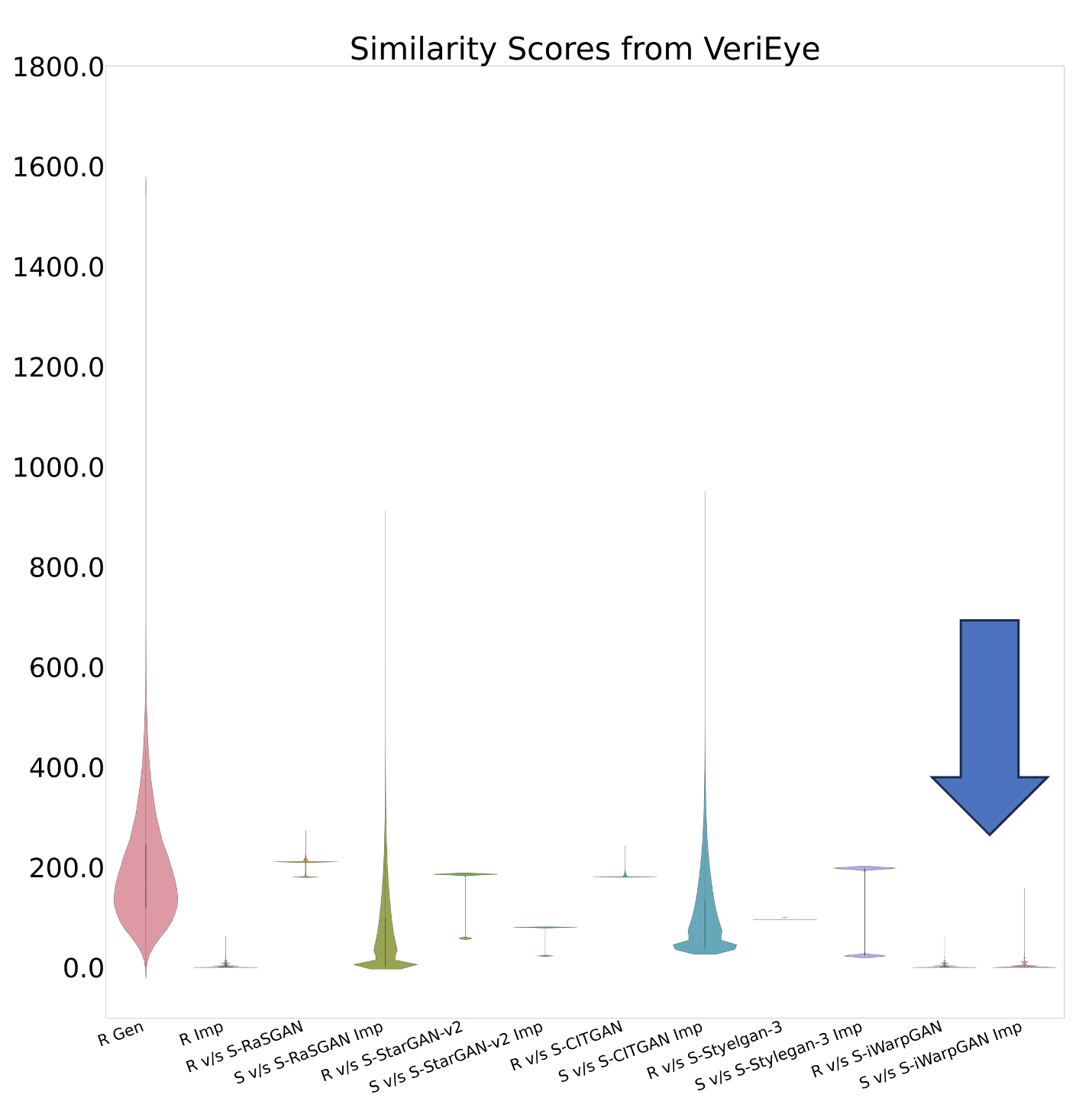}
    \caption{This figure shows the uniqueness of iris images generated using iWarpGAN when the GANs are trained using CASIA-Iris-Thousand dataset. The y-axis represents the similarity scores obtained using VeriEye. Here, R=Real, S=Synthetic, Gen=Genuine and Imp=Impostor. As indicated by the arrow here, similarity between images generated by iWarpGAN is the lowest in comparison to other GANs, followed by StyleGAN-3. This indicates the uniqueness of the identity generated by these GANs with respect to real identities and also the generated identities.}
    \vspace{-2mm}
    \label{fig:Vplot-Casia1000.png}
\end{figure*}

\begin{figure*}
    \centering
    \includegraphics[width=0.85\textwidth]{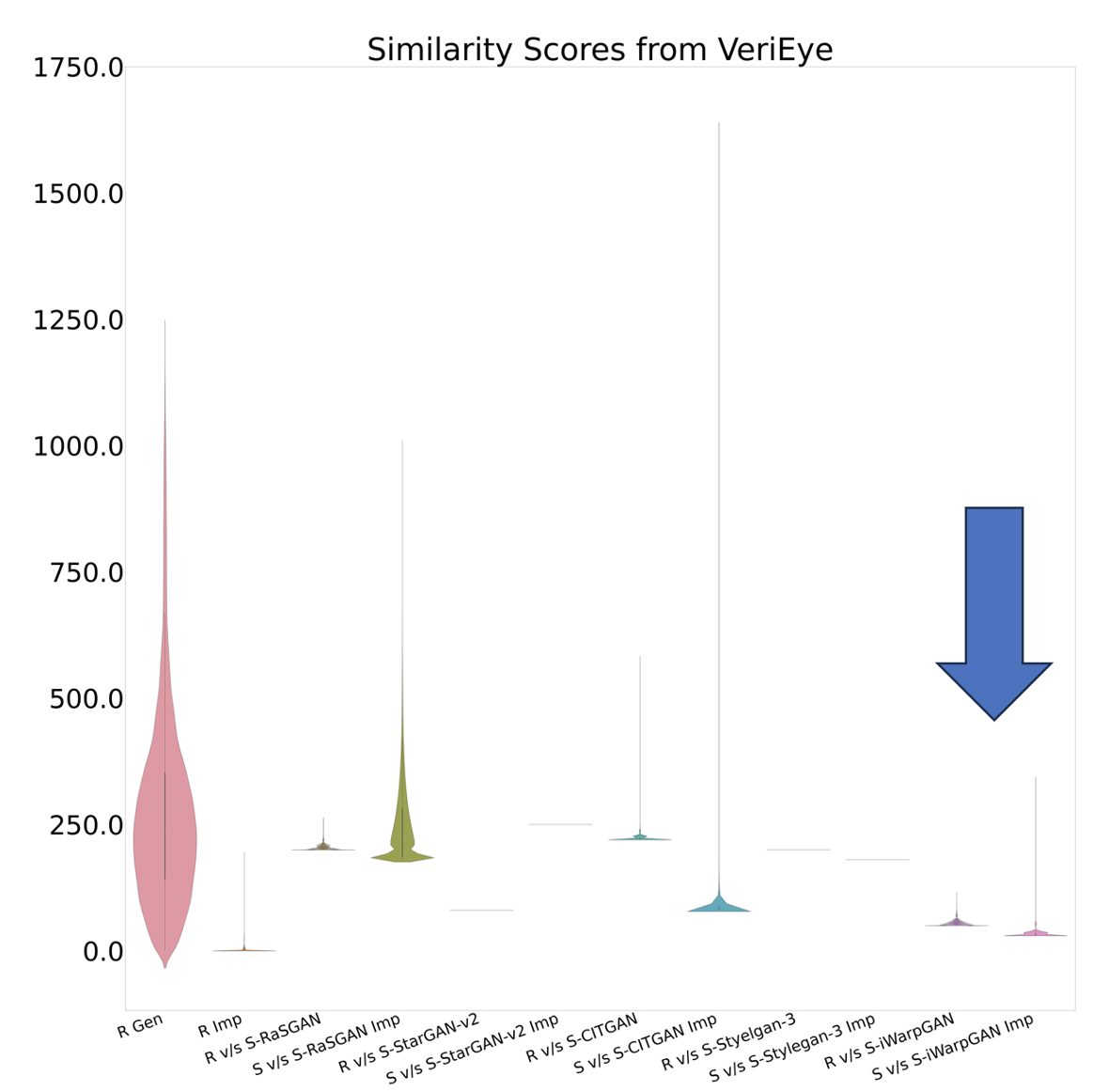}
    \caption{This figure shows the uniqueness of iris images generated using iWarpGAN when the GANs are trained using CASIA-CS iris dataset. The y-axis represents the similarity scores obtained using VeriEye. Here, R=Real, S=Synthetic, Gen=Genuine and Imp=Impostor. As indicated by the arrow here, similarity between images generated by iWarpGAN is the lowest in comparison to other GANs, followed by StyleGAN-3. This indicates the uniqueness of the identity generated by these GANs with respect to real identities and also the generated identities.}
    \vspace{-2mm}
    \label{fig:Vplot-Casia-CS.png}
\end{figure*}

\begin{figure*}
    \centering
    \includegraphics[width=0.85\textwidth]{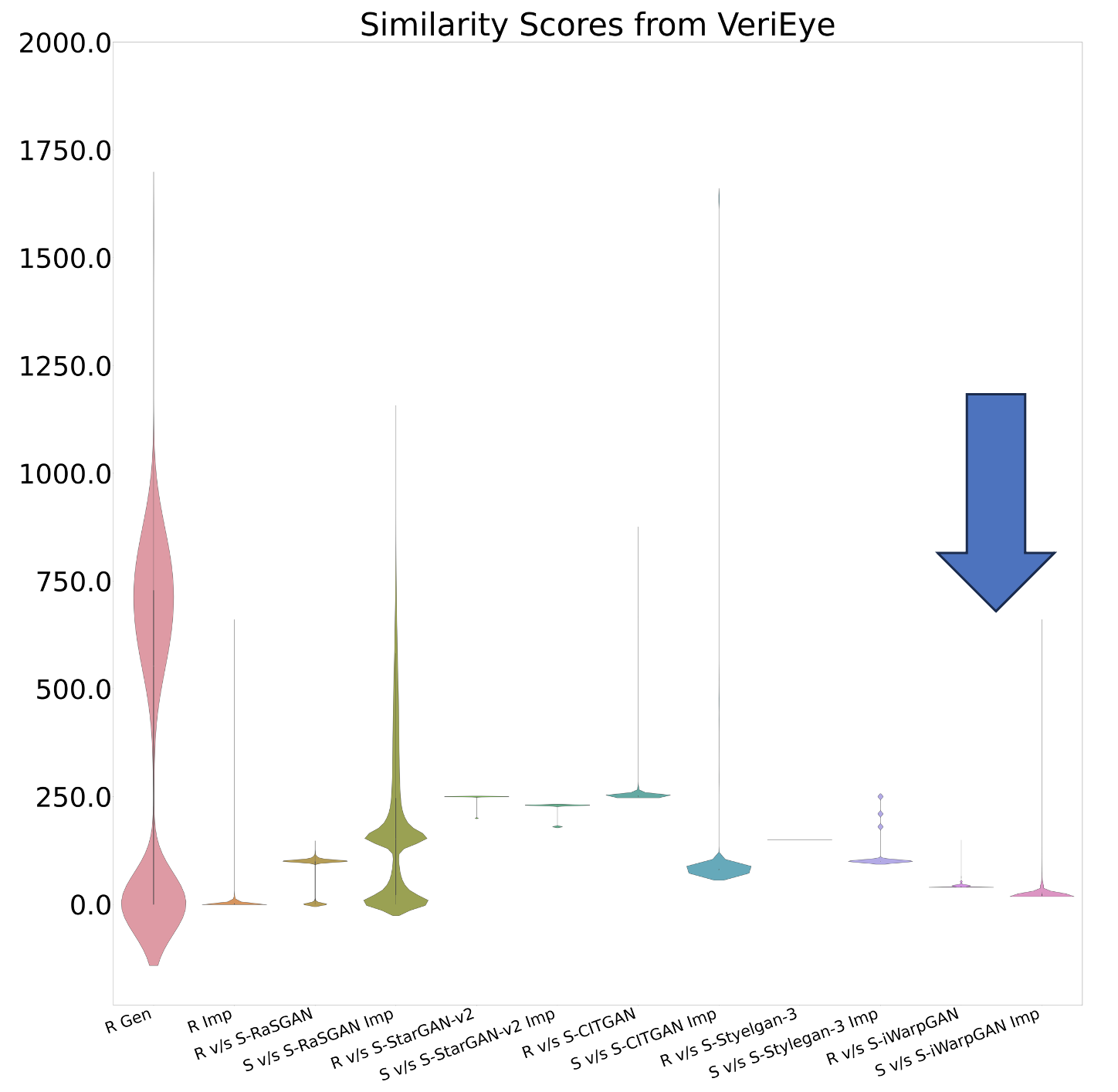}
    \caption{This figure shows the uniqueness of iris images generated using iWarpGAN when the GANs are trained using IITD iris dataset. The y-axis represents the similarity scores obtained using VeriEye. Here, R=Real, S=Synthetic, Gen=Genuine and Imp=Impostor. As indicated by the arrow here, similarity between images generated by iWarpGAN is the lowest in comparison to other GANs, followed by StyleGAN-3. This indicates the uniqueness of the identity generated by these GANs with respect to real identities and also the generated identities.}
    \vspace{-5mm}
    \label{fig:Vplot-IITD.png}
\end{figure*}

Analysis: The above-mentioned experiments replicate the experimental protocols defined in \cite{yadav2023} to evaluate the inter and intra-class variations in the generated iris dataset. It also analyzes the distinction of the generated identities from the real irides in the training dataset. As depicted in Figures \ref{fig:Vplot-Casia1000.png}, \ref{fig:Vplot-Casia-CS.png}, and \ref{fig:Vplot-IITD.png}, the iris images produced by iWarpGAN do not exhibit a high degree of resemblance to the real irides from training set, unlike the outcomes from other GAN methods. This indicates iWarpGAN's proficiency in generating iris patterns with identities that diverge from those found in the training set. Moreover, by examining the overlap between the imposter distribution of the synthetically generated iris images and that of the genuine iris images, it becomes evident that the generated identities are distinct from one another. Hence, we can conclude that iWarpGAN has the capability to generate fully synthetic irides images with identities that are significantly different from irides in training set, while other GANs are efficient in generating only partially synthetic irides.

\begin{table*}[t]
\centering
\caption{PAD-Experiment-0: True Detection Rate (TDR in \%) at three different False Detection Rate (FDR) of different iris PAD methods in baseline experiment (PAD-Experiment-0) where PA detectors are trained using real bonafide and PA samples. The PA samples used in this experiment are imbalanced across different PA classes \cite{yadav2021}.}
\label{tab:PAD-Exp0}
\scalebox{1.0}{%
\begin{tabular}{@{}lccccc@{}}
\toprule
      & \multicolumn{1}{l}{\textbf{BSIF+SVM} \cite{doyle2015}} & \multicolumn{1}{l}{\textbf{Fine-Tuned VGG-16} \cite{gatys2015}} &
      \multicolumn{1}{l}{\textbf{Fine-Tuned AlexNet} \cite{alex2012}} &
      \multicolumn{1}{l}{\textbf{D-NetPAD} \cite{sharma2020}}  \\ \midrule
TDR (@0.1\%)  & 3.32   & 85.25   & 86.10   & 87.94  
\\
TDR (@0.2\%)  & 6.15   & 83.86   & 87.29   & 88.91    
\\
TDR (@1.0\%)    & 28.11  & 89.07   & 90.51  & 92.54   
\\ \bottomrule
\end{tabular}}
\end{table*}

\begin{table*}[t]
\centering
\caption{PAD-Experiment-1: True Detection Rate (TDR in \%) at 1\% False Detection Rate (FDR) of different iris PAD methods when trained using real bonafide irides, real PAs and synthetic PAs generated using different GAN methods. Comparing the baseline performance with the performance here, we can see that iris PA images generated by StyleGAN-3 and iWarpGAN are better at replacing the real PAs.}
\label{tab:PAD-Exp1}
\scalebox{1.0}{%
\begin{tabular}{@{}lccccc@{}}
\toprule
      & \multicolumn{1}{l}{\textbf{BSIF+SVM} \cite{doyle2015}} & \multicolumn{1}{l}{\textbf{Fine-Tuned VGG-16} \cite{gatys2015}} &
      \multicolumn{1}{l}{\textbf{Fine-Tuned AlexNet} \cite{alex2012}} &
      \multicolumn{1}{l}{\textbf{D-NetPAD} \cite{sharma2020}}  \\ \midrule
RaSGAN  & 28.31   & 81.11   & 86.74   & 87.97  
\\
CIT-GAN  & 29.43   & 85.81   & 88.37   & 88.86  
\\
StarGAN-v2    & 29.73  & 83.47   & 88.71   & 88.45   
\\ 
StyleGAN-3    & 34.05   & 88.14   & 90.95   & 90.75   
\\ 
iWarpGAN    & 32.09  & 86.58   & 89.18   &  90.28
\\ 
\bottomrule
\end{tabular}}
\end{table*}

\begin{table*}[t]
\centering
\caption{PAD-Experiment-2: True Detection Rate (TDR in \%) at 1\% False Detection Rate (FDR) of different iris PAD methods when they are trained using real bonafide irides alongside a balanced collection of PA samples using both real and synthetic irides. The performance here shows the utility of generated irides in improving the performance iris PAD methods.}
\label{tab:PAD-Exp2}
\scalebox{1.0}{%
\begin{tabular}{@{}lccccc@{}}
\toprule
      & \multicolumn{1}{l}{\textbf{BSIF+SVM} \cite{doyle2015}} & \multicolumn{1}{l}{\textbf{Fine-Tuned VGG-16} \cite{gatys2015}} &
      \multicolumn{1}{l}{\textbf{Fine-Tuned AlexNet} \cite{alex2012}} &
      \multicolumn{1}{l}{\textbf{D-NetPAD} \cite{sharma2020}}  \\ \midrule
RaSGAN  & 50.52    & 88.62    & 88.15  & 94.90  
\\
CIT-GAN  & 51.11   & 91.60  & 92.70   & 97.89
\\
StarGAN-v2  & 50.69   & 92.39   & 88.38  & 95.59    
\\ 
StyleGAN-3  & 56.14   & 95.02   & 93.38  & 98.39   
\\ 
iWarpGAN    & 55.25   & 93.69   & 92.99  & 98.20
\\ 
\bottomrule
\end{tabular}}
\end{table*}


\subsection{Utility of Synthetically Generated Irides}
The experiments in this section evaluate the usefulness of the synthetically generated irides in the task of presentation attack detection as well as iris recognition.

\subsubsection{Presentation Attack Detection}
As discussed earlier, lack of sufficient number and variations of PA samples can affect the generalizability of different PA detection methods, especially those based on deep networks that need large number of training samples for better performance. Therefore, we outline experimental protocol that was established to assess the utility of the synthetically generated iris PAs. To generate the synthetic data (for both bonafide and PAs), the GANs are trained using 14,970 bonafide iris images, 6,016 printed eyes, 4,014 cosmetic contact lenses and 276 artificial eyes from the iris PA datasets mentioned earlier. Note that CIT-GAN, StarGAN-v2, StyleGAN-3 and iWarpGAN can handle style transfer from one domain to another, i.e., image generation using these GANs can extend to multiple domains/style. However, RaSGAN can facilitate only single domain image generation. Therefore, multiple RaSGAN networks are trained to be able to generate different types of PAs as well as bonafides.

We analyzed the efficacy of several iris Presentation Attack Detection (PAD) techniques, including VGG-16 \cite{gatys2015}, BSIF \cite{doyle2015}, D-NetPAD \cite{sharma2020}, and AlexNet \cite{alex2012}, within these different experimental configurations for analytical scrutiny. It is noteworthy that D-NetPAD has been recognized as one of the top-performing PAD algorithms, particularly highlighted for its performance in the iris liveness detection challenge (LivDet-20 edition).
\vspace{2mm}

PAD-Experiment-0: In this baseline experiment, we demonstrate the performance of different iris PAD methods on iris PA datasets mentioned in Dataset section. The PAD methods are trained with 14,970 bonafide iris images and 10,306 instances of PAs, which include 4,014 cosmetic contact lenses, 276 artificial eyes and 6,016 printed eyes. For testing, the dataset contain 9,439 bonafide iris images alongside 9,896 PA instances, which broke down into 2,720 cosmetic contacts, 404 artificial eyes and 6,772 printed eyes.
\vspace{2mm}

PAD-Experiment-1: Here, we aim to evaluate the realism of synthetically generated PAs against real PAs by training the iris PA detection methods with 14,970 bonafide samples and synthetically generated 6,016 printed eyes, 276 artificial eyes and 4,014 cosmetic contact lenses. For testing, the set contains 9,439 bonafide iris images alongside 9,896 PA instances, that consists of 2,720 cosmetic contacts, 404 artificial eyes and 6,772 printed eyes.
\vspace{2mm}

PAD-Experiment-2: Here, we aim to evaluate the usefulness of the generated PA sampled for balanced training when imbalanced PA classes are improved by adding synthetically generated PAs to the training. Therefore, in this experiment, PAD methods are trained using 14,970 bonafide irides alongside a balanced collection of 15,000 PA samples. This assembly comprises of 276 artificial eyes, 4,014 cosmetic contact lenses, and 5,000 printed eyes which are real PAs; in addition, 4,724 synthetic artificial eyes and 986 synthetic cosmetic contact lenses are utilized in the training set. Similar to the previous experiment, testing is done on 9,439 bonafide iris images alongside 9,896 PA instances that consists of 2,720 cosmetic contacts, 404 artificial eyes and 6,772 printed eyes.
\vspace{2mm}


Analysis: The variation in the number of samples across different PA categories influences the effectiveness of the PAD techniques. This impact is evident when the outcomes of PAD-Experiment-0 are compared with PAD-Experiment-1 \& 2. In PAD-Experiment-2, the PAD methods are trained with 9,439 bonafide samples and an equalized set of PA samples (namely, 5,000 from each PA category), including both real and synthesized PAs. Based on the data in Table \ref{tab:PAD-Exp0} and Table \ref{tab:PAD-Exp2}, there is a discernible enhancement in the performance of each PAD approach when trained with balanced samples from each class. Moreover, a comparative analysis of synthetic PA samples and actual PA samples was conducted through PAD-Experiment-1. In this experiment, a portion of the real PA samples in the training set was substituted with synthetic PAs. When comparing the performance metrics in Table \ref{tab:PAD-Exp0} and Table \ref{tab:PAD-Exp1}, only a marginal difference in PAD efficacy is noticeable.

\subsubsection{Iris Recognition}
As mentioned earlier, the lack of sufficient number of unique identities in the dataset with large intra-class variations can affect the training and testing of many iris recognition methods, especially recognition methods based on deep networks that needs large number of training samples for better performance. As seen from the previous experiments, among the GANs studied in this paper, only iWarpGAN has the capability of generating irides whose identities are sufficeintly  different from that of the training data. Therefore, we train iWarpGAN using the CASIA-Iris-Thousand, CASIA-CSIR and IIT-Delhi iris datasets, separately, to generate synthetic irides with both inter and intra-class variations. The generated dataset is utilized in this experiment to evaluate its usefulness for improved iris recognition.

Recog-Experiment-0: In this baseline experiment, EfficientNet \cite{hsiao2021}, ResNet-101 \cite{minaee2019} and DenseNet-201 are trained using the triplet training approach. Training and testing has been done using cross-dataset method, i.e., when trained using real irides from CASIA-Iris-Thousand and CASIA-CSIR, testing is done on IITD-iris dataset.
\vspace{2mm}

Recog-Experiment-1: This experiment focuses on assessing the impact of synthetic iris dataset on enhancing the performance of iris recognition methods based on deep learning. In this context, EfficientNet, ResNet-101 and DenseNet-201 are trained with not only the real irides from CASIA-Iris-Thousand, CASIA-CSIR, and IITD-iris datasets but also with a synthetically generated iris dataset derived from iWarpGAN.
\vspace{2mm}

\begin{figure*}
    \centering
    \captionsetup{justification=centering}
    \includegraphics[width=0.95\textwidth]{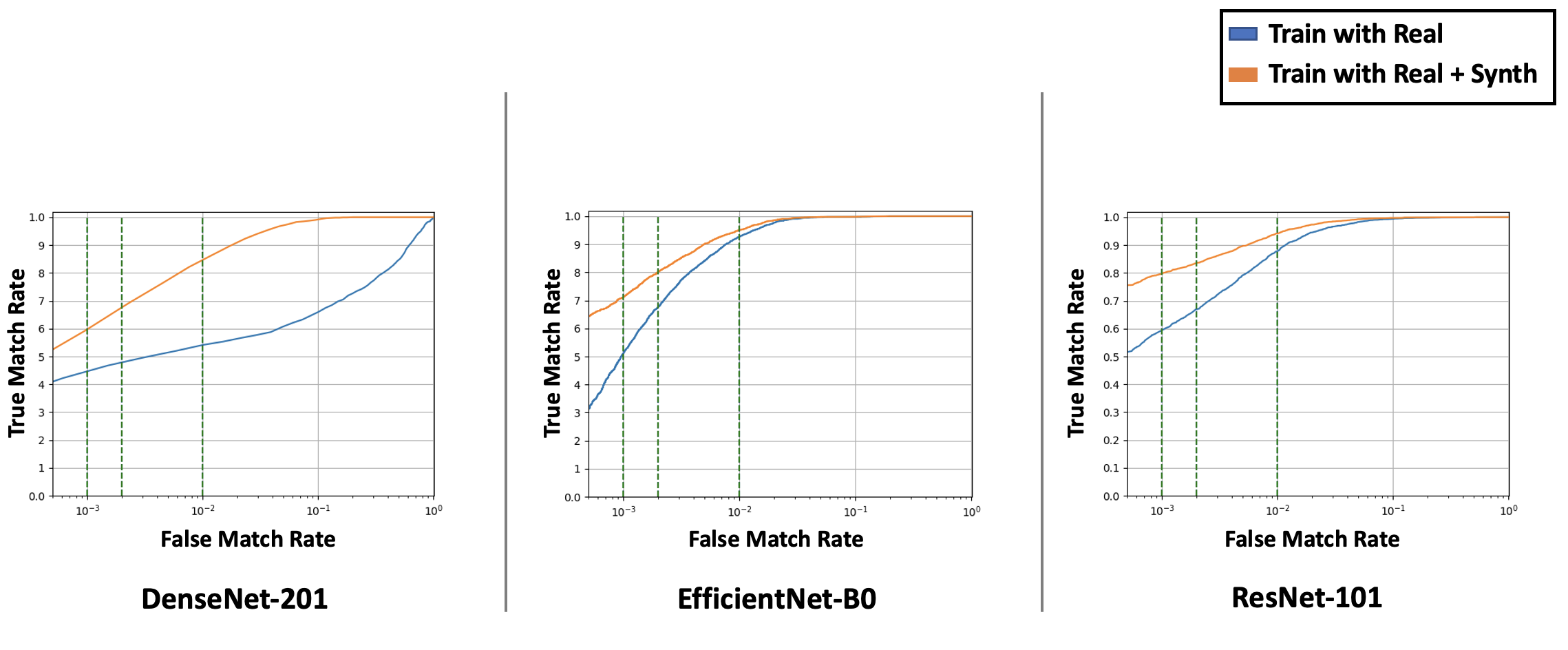}
    \caption{This figure shows the performance of DenseNet-201, EfficientNet and ResNet-101 in the cross-dataset evaluation scenario i.e., trained using CASIA-Iris-Thousand \& CASIA-CSIR datasets and tested using IIT-Delhi iris dataset. Improvement in the performance is seen when size of training set is increased using synthetic irides.}
    \vspace{-5mm}
    \label{fig:ROC-IITD}
\end{figure*}

\begin{figure*}
    \centering
    \captionsetup{justification=centering}
    \includegraphics[width=0.95\textwidth]{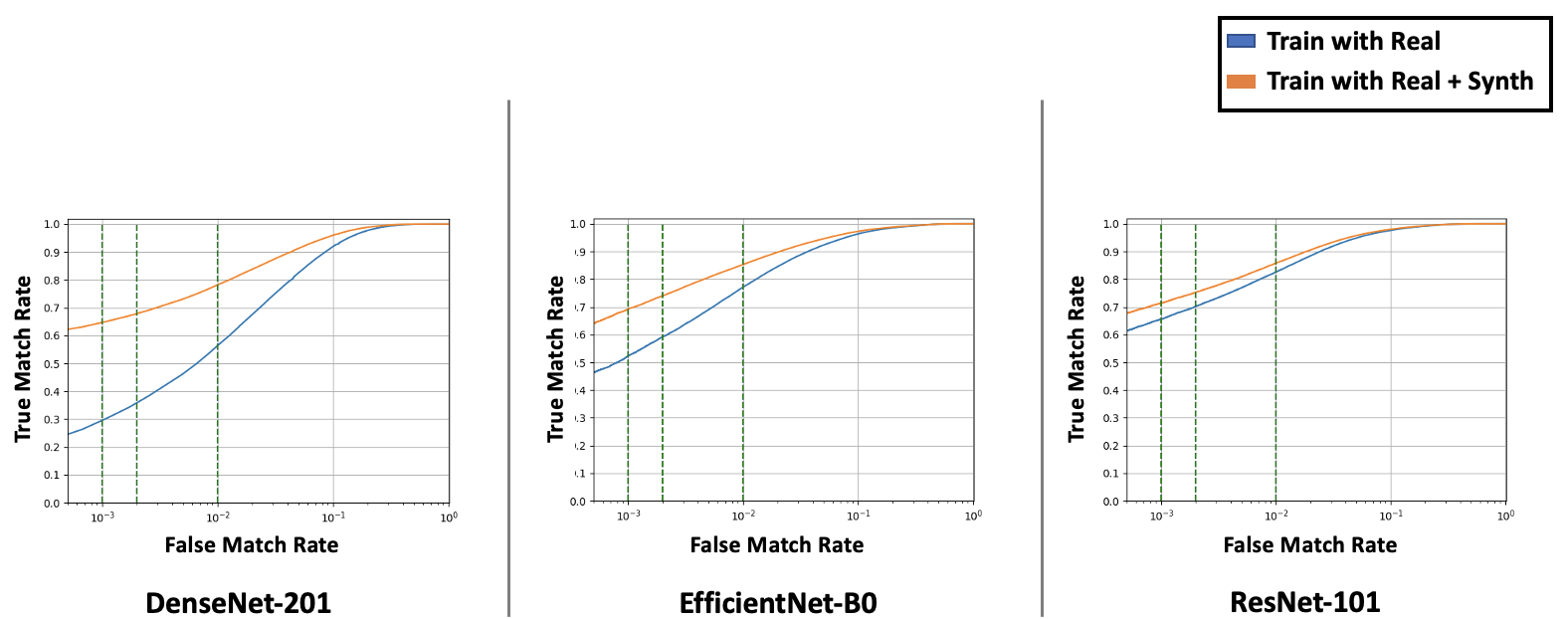}
    \caption{This figure shows the performance of DenseNet-201, EfficientNet and ResNet-101 in the cross-dataset evaluation scenario i.e., trained using CASIA-Iris-Thousand \& IIT-Delhi iris datasets and tested using CASIA-CSIR dataset. Improvement in the performance is seen when size of training set is increased using synthetic irides.}
    \vspace{-5mm}
    \label{fig:ROC-Casia-CS}
\end{figure*}

Analysis: Figures \ref{fig:ROC-IITD} and \ref{fig:ROC-Casia-CS} illustrate that the recognition accuracy of iris, matchers based on deep learning, is enhanced with the incorporation of a larger dataset  with synthetic samples. This enhancement is particularly evident when the matcher is trained using a combination of both real iris images and those synthetically produced by iWarpGAN. Although the baseline performance of ResNet-101 and EfficientNet is somewhat modest, they exhibits a notably substantial enhancement in the performance. Similar, behavior is seen for DenseNet-201.

\section{Summary \& Future Work}
In this section, we summarize the studies conducted in this paper and also discuss the future scope in the field of generating synthetic irides.

\subsection{Current Techniques \& their Limitations}
In this research, we reviewed and analyzed different GAN methods to generate synthetic images for both bonafide irides and different presentation attack instruments. The generated irides were evaluated for their realism, quality, unique identities and utility. Using these experiments as our criteria for comparison, we conclude that: (1) GAN methods like RaSGAN, StarGAN-v2 and CIT-GAN can generate realistic looking synthetic dataset, but fail to generate enough samples whose identities are different from those in the training dataset, i.e., the identities in the generated dataset has high similarity with the training dataset and also with itself. Similar behavior was seen for StyleGAN-3, however the images generated by StyleGAN-3 are highly realistic and very close to original dataset in terms or quality (as seen in Figure \ref{fig:ISO-Exp}). On the other hand, iWarpGAN showed the capability of generating synthetic irides that has both inter and intra-class variations, which can augment real iris datasets for training and testing iris recognition methods. This method is scalable to multiple domains (using attribute vector), and can also be utilized to generate bonafide and various PAs that can be used to enhance the performance of different PAD methods (as shown in \ref{tab:PAD-Exp2}). While this method provides solutions for both fully and partially synthetic iris generation, iWarpGAN utilizes image transformation, whereby the network requires both an input and a style reference image to modify the identity and style, resulting in the generation of an output image. Such a process could potentially constrain the range of features that iWarpGAN is able to explore. Also, there is still some similarity observed between training samples and generated irides.

\subsection{Future Work \& Scope}
Numerous researchers are dedicating their efforts to the creation of synthetic face images encompassing varied attributes, styles, identities, spectra, and more \cite{osadchy2017, melzi2023, queiroz2010, kolf2023}. However, more study has to be done in the field of synthetic iris generation. This opens up a wide array of opportunities that warrant in-depth investigation and exploration. Some of the possible future work in this field are listed here:

\begin{itemize}
    
    \item \textbf{Generalizable Solution for Fully-Synthetic Iris Images}: The first area of exploration involves developing a generalizable solution for creating fully synthetic iris images. This involves not just replicating the physical appearance of an iris but also ensuring that the synthetic images can adapt or respond to different lighting conditions and camera specifications, just as a real iris would. Such a solution would have huge implications for enhancing the realism and applicability of synthetic irides in various fields, including iris recognition and presentation attack detection. Also, demographic attributes such as gender, age, etc. needs to be accounted for while generating synthetic iris images.
    
    \item \textbf{Generating Ocular Images}: Another intriguing direction is the generation of complete ocular images, which include not only the iris but also other parts of the eye. So far, the research work in this field manly focuses on generating cropped iris images and for some cases the image quality deteriorates as more information is introduced in the image  \cite{yadav2020}. Therefore, this area needs attention from researchers to be able to study the other distinguishing features of an eye apart from irides \cite{cardoso2013iris, kaur2020}. Creating realistic ocular images that accurately represent the myriad variations in human eyes could also aid in the development of more robust facial recognition technologies by providing a method to generate faces that also captures the intricate details of a real iris, which seems to be missing from most of the face generation methods.
    
    \item \textbf{Synthetic Iris Videos to Mimic Liveness of Real Irides}: The creation of synthetic iris videos that can mimic the liveness of real irides is a particularly challenging topic. Such advancements would be beneficial in developing more robust PA detection methods. By simulating the natural movements and minute dynamic changes in the iris, these videos could provide an authentic and effective tool for training and improving liveness detection algorithms in iris recognition systems. Also as mentioned earlier, this could also aid in developing a robust facial recognition technologies.   
    
    \item \textbf{Multi-spectrum Iris Image Generation}: The generation of multi-spectrum iris images presents another frontier \cite{ross2009, boyce2006}. The human iris exhibits different characteristics under various light spectrum - a feature that is often leveraged in biometric systems. Developing synthetic iris images that can accurately reflect these multi-spectral properties would not only enhance the realism of these images but also expand their utility in biometric recognition systems. Such multi-spectrum images could serve as a valuable resource for researchers and developers, offering a versatile tool for testing and improving multi-spectral iris recognition technologies.
    
    \item \textbf{Improved GAN Latent Space Interpretability:} As indicated by iWarpGAN, enhancing GANs to produce disentangled representations in their latent space is vital for generating full synthetic iris images. Here, improving semantic interpretability can help ensure that changes in the latent space correspond to meaningful and coherent changes in the generated output \cite{kahng2019, bau2018}. Also, developing new visualization techniques to map and understand the high-dimensional latent space is another crucial step.
    
    \item \textbf{Diffusion Diffusion Generative Adversarial Networks:} Diffusion GANs \cite{wang2022diffusion} offer a promising approach for generating synthetic iris images with high realism with respect to real iris images. By employing a diffusion process, these networks iteratively refine generated images over multiple steps, allowing for better control and coherence in synthesis. With this, Diffusion GANs can capture complex spatial dependencies inherent in iris textures, ensuring the production of realistic patterns consistent with real iris images. Moreover, they exhibit improved stability and convergence during training, mitigating issues like mode collapse and artifact generation.

    \item \textbf{Application in Deepfake detection:} Synthetic images can play a crucial role in deepfake detection \cite{rana2022,zhao2021} by serving as a tool for training and evaluating detection algorithms. By incorporating a diverse range of synthetic data into the training process, detection algorithms can better generalize and identify subtle inconsistencies or artifacts indicative of deep fakes. Moreover, synthetic images allow researchers to experiment with various image editing techniques, enhancing the robustness of detection systems against emerging threats. Therefore, synthetic images can serve as a critical resource in advancement and development of robust deep fake detection methods.
\end{itemize}

The potential applications of successfully generated synthetic iris images are vast and varied. In security and biometric recognition systems, these images can help improve the accuracy and robustness of systems by providing a diverse range of data for training and testing. In the medical field, synthetic iris images could be used for training purposes, enabling medical professionals to recognize and diagnose eye-related diseases more effectively.

Furthermore, in the realm of entertainment and virtual reality, realistic synthetic iris images could enhance the visual experience by providing more lifelike and expressive characters. The ability to generate eyes that accurately mimic human emotions could revolutionize the way we interact with virtual environments and characters. In conclusion, while the generation of realistic and unique synthetic iris images is still in the development stage, it presents an opportunity for research and exploration.

\section{Acknowledgments}
This research is based upon work supported by by NSF CITeR funding. The views and conclusions contained herein are those of the authors and should not be interpreted as necessarily representing the official policies, either expressed or implied by NSF CITeR.

\balance
{\small
\bibliographystyle{ieee}
\bibliography{egbib}
}

\end{document}